\documentclass[10pt,journal,compsoc]{IEEEtran}
\usepackage{graphicx}           
\usepackage{amsthm}
\usepackage{amsmath,amsfonts, empheq, mathabx}
\usepackage{times,epsfig}
\usepackage{psfrag}
\usepackage{stfloats}
\usepackage{amssymb}
\usepackage{multirow}
\usepackage{algorithmic}
\usepackage[ruled,vlined,linesnumbered]{algorithm2e}
\usepackage{cite}
\usepackage{CJK}
\usepackage{url}
\usepackage{array}
\usepackage{soul}
\usepackage{booktabs,ragged2e, tabularx}
\usepackage[flushleft]{threeparttable}
\usepackage{makecell}
\usepackage{bm}
\usepackage{color}
\usepackage{hyperref}
\usepackage[dvipsnames]{xcolor}

\theoremstyle{definition}
\newtheorem{thm}{Theorem}
\newtheorem{lem}{Lemma}

\makeatletter
\newenvironment{proof*}[1][\proofname]{\par
  \pushQED{\qed}%
  \normalfont \partopsep=\z@skip \topsep=\z@skip
  \trivlist
  \item[\hskip\labelsep
        \itshape
    #1\@addpunct{.}]\ignorespaces
}{%
  \popQED\endtrivlist\@endpefalse
}
\makeatother

\makeatletter
\def\thm@space@setup{\thm@preskip=1.5pt
\thm@postskip=1.5pt}
\makeatother

\begin{document}
\title{Filling the Missing: Exploring Generative AI for Enhanced Federated Learning over Heterogeneous Mobile Edge Devices\vspace{-5pt}
}
\author{
Peichun~Li, Hanwen~Zhang, 
Yuan~Wu,~\IEEEmembership{Senior~Member,~IEEE},
Liping~Qian,~\IEEEmembership{Senior~Member,~IEEE},
Rong~Yu,~\IEEEmembership{Member,~IEEE},
Dusit Niyato,~\IEEEmembership{Fellow,~IEEE},
and Xuemin (Sherman) Shen,~\IEEEmembership{Fellow,~IEEE} \vspace{-5pt}
\IEEEcompsocitemizethanks{
\IEEEcompsocthanksitem P. Li, H. Zhang, and Y. Wu are with the State Key Laboratory of Internet of Things for Smart City, University of Macau, Macau, China, and also with Department of Computer and Information Science, University of Macau, Macau, China. (e-mail: peichunli@um.edu.mo,  hw.zhang@connect.um.edu.mo, yuanwu@um.edu.mo).
\IEEEcompsocthanksitem L. Qian is with College of Information Engineering, Zhejiang University of Technology, Hangzhou 310023, China (email:lpqian@zjut.edu.cn).
\IEEEcompsocthanksitem R. Yu is with School of Automation, Guangdong University of Technology, Guangzhou, China, and also with Guangdong Key Laboratory of IoT Information Technology, Guangzhou, China. (e-mail: yurong@ieee.org).
\IEEEcompsocthanksitem D. Niyato is with the School of Computer Science and Engineering, Nanyang Technological University, Singapore, Block N4-02a-32, Nanyang Avenue, Singapore. (e-mail: dniyato@ntu.edu.sg).
\IEEEcompsocthanksitem X. Shen is with the Department of Electrical and Computer Engineering, University of Waterloo, Waterloo, ON N2L 3G1, Canada (e-mail: sshen@uwaterloo.ca).
\IEEEcompsocthanksitem \textit{Yuan Wu is the corresponding author.}

}
}

\IEEEtitleabstractindextext{%
\begin{abstract}
Distributed Artificial Intelligence (AI) model training over mobile edge networks encounters significant challenges due to the data and resource heterogeneity of edge devices. The former hampers the convergence rate of the global model, while the latter diminishes the devices' resource utilization efficiency. In this paper, we propose a generative AI-empowered federated learning to address these challenges by leveraging the idea of FIlling the MIssing (FIMI) portion of local data. Specifically, FIMI can be considered as a resource-aware data augmentation method that effectively mitigates the data heterogeneity while ensuring efficient FL training. We first quantify the relationship between the training data amount and the learning performance. We then study the FIMI optimization problem with the objective of minimizing the device-side overall energy consumption subject to required learning performance constraints. The decomposition-based analysis and the cross-entropy searching method are leveraged to derive the solution, where each device is assigned suitable AI-synthesized data and resource utilization policy. Experiment results demonstrate that FIMI can save up to 50\% of the device-side energy to achieve the target global test accuracy in comparison with the existing methods. Meanwhile, FIMI can significantly enhance the converged global accuracy under the non-independently-and-identically distribution (non-IID) data.
\end{abstract}
\vspace{-6pt}
}

\maketitle

\section{Introduction}
Generative artificial intelligence (AI) has advanced significantly in content creation. Notably, vision models such as stable diffusion and Imagen have demonstrated their impressive capability in generating photo-realistic images \cite{rombach2022high, saharia2022photorealistic, xu2023unleashing}. These advancements have sparked new possibilities that generative models can play a significant role in enhancing the centralized training process. Specifically, the synthesized images produced by generative models can be effectively utilized to improve the performance of classification models \cite{azizi2023synthetic}. 
In the realm of federated learning (FL), where isolated devices with limited storage collaborate in training a global model, data and resource heterogeneity have emerged as crucial challenges \cite{mcmahan2016comm, wu2022ai, chen2020wireless, yu2021toward, yang2022federated}. These factors lead to a slower convergence rate during the global training process. Consequently, an intriguing and pertinent question arises, i.e., can generative AI be harnessed to address these issues and improve the performance of FL ultimately?

\begin{figure}[t]\centering
  \includegraphics[width=0.48\textwidth]{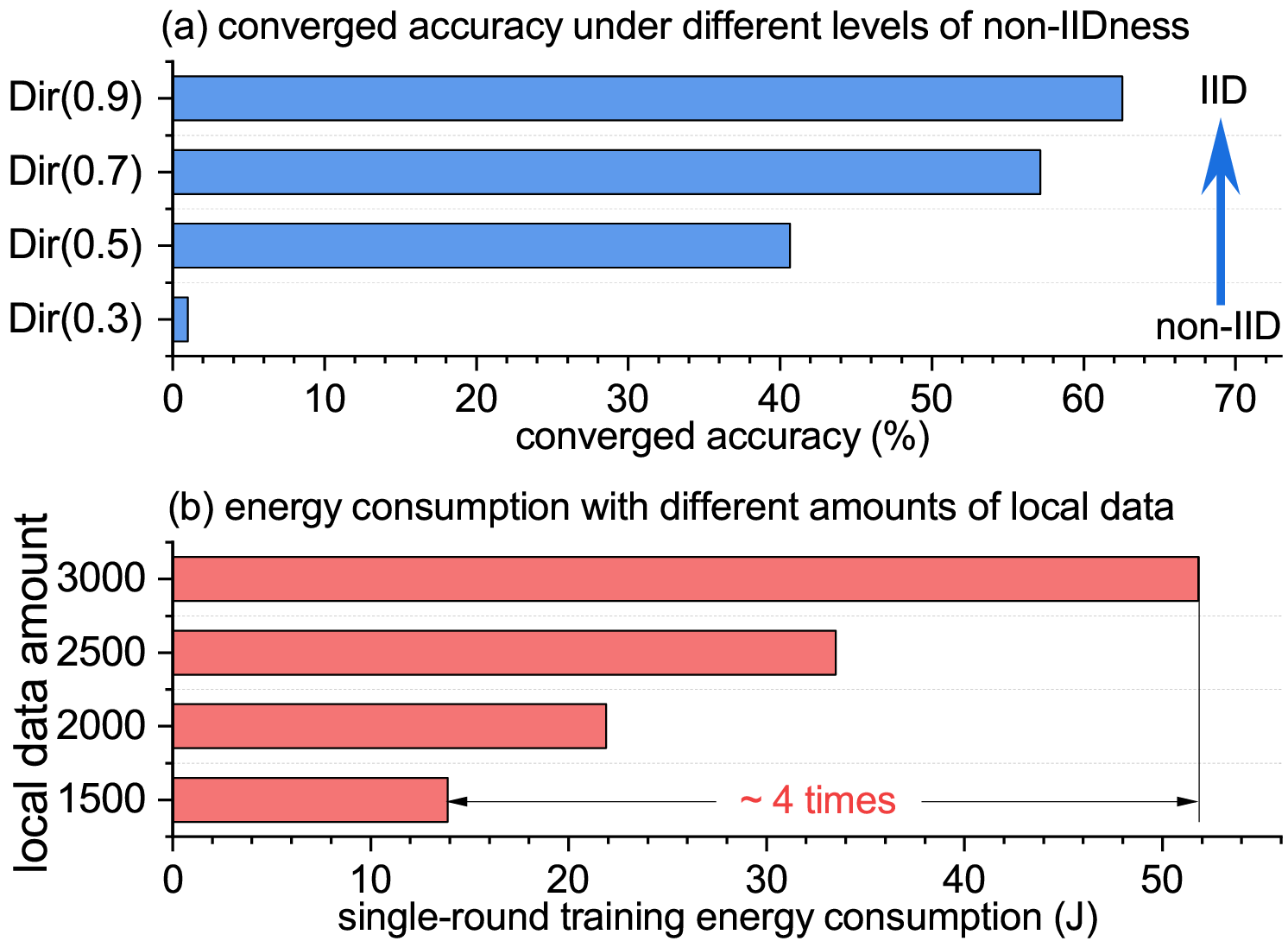}
\caption{The impacts of data distribution and data amount on the training performance.} \label{fig:intro}
\vspace{-8pt}
\end{figure}

In this paper, we first investigate how the distribution of federated local data impacts the training performance of FL. Specifically, we focus on the VGG9 model \cite{simonyan2014very} with 20 Nvidia Jetson Nano devices on CIFAR100 dataset \cite{krizhevsky2009learning}. We regulate the local dataset into different degrees of Dirichlet distribution, leading to varying levels of non-independently-and-identically distribution (non-IID) data \cite{wang2020federated, zhang2022scalable}. As shown in the top subplot of Fig. \ref{fig:intro}, we observe that a well-distributed Dir(0.9) significantly improves convergence accuracy compared to Dir(0.3). Here, Dir($z$) means the Dirichlet distribution with parameter $z$, and a larger $z$ means a more homogeneous distribution of the samples (i.e., less non-IID distribution). The observation indicates that augmenting the data to eliminate the non-IID feature of local data holds promise for enhancing the learning performance of FL. However, the augmented training data introduces extra computation costs for edge devices. As shown in the bottom subplot of Fig.~\ref{fig:intro}, doubling the amount of data results in about a fourfold increase in energy consumption under a fixed training latency.

The observations above offer insights for developing an efficient and effective generative AI-enhanced FL system. To address data heterogeneity, it is recommended to leverage generative AI to compensate for the missing portion of local data. This strategy can bridge the gaps in data distribution across devices and improve convergence accuracy. However, it is crucial to take into account the locally available resources and adapt the amount of synthesized data on different devices accordingly. The majority of current literature on generative AI-enhanced training methods tends to prioritize improving the model accuracy in a centralized training manner, disregarding the computation cost induced by the augmented training dataset \cite{zheng2023toward, azizi2023synthetic}. These approaches may shorten the battery lifetime and potentially degrade the user experience when integrated into the FL system directly. Therefore, we propose a resource-aware data augmentation framework, called FIlling the Missing (FIMI), which is designed for generative AI-enhanced FL systems. 

The key idea behind FIMI is to leverage the concept of filling the missing portion of data for each device. Our goal is to improve the learning performance of FL by utilizing a pre-trained generative AI model to synthesize customized local data while ensuring the training efficiency of edge devices.
To this end, we first outline the generic workflow of our generative AI-empowered FL system. We then empirically measure the relationship between the amount of training data and model learning error to quantitatively analyze the learning performance of FL. Furthermore, we propose an optimization problem for minimizing the device-side energy subject to the learning performance constraints. To solve this formulated problem, we introduce auxiliary variables and decompose the main problem into two sub-problems along with a top-layer searching problem. To find solutions for the sub-problems, the convex optimization methods are leveraged to obtain the optimal solution based on the theoretical analysis. For the top-layer problem, the continuous cross-entropy algorithm is employed to efficiently search the solution. 
After that, a data entropy maximization problem for mitigating the non-IID feature of local data is conducted to obtain the category-wise training sample amount on each device.
In this way, each wireless edge device is assigned a personalized solution, where the data synthesis strategy and resource utilization policy are optimized to minimize the device-side energy consumption while improving the learning performance.

Our main contributions are summarized as follows.
\begin{itemize}
	\item We propose a generative AI-enhanced FL framework called FIMI to leverage the pre-trained generative models for filling the missing portion of the local data. The calibrated distribution of local data improves the convergence rate of FL training.
	\item By combining the theoretical and empirical analysis, we quantitatively establish the relationship between the training data amount and the learning performance of FL.
    \item We formulate a joint optimization of data synthesis strategy and resource utilization policy, with the objective of minimizing the energy consumption of FIMI over wireless networks. We also propose an efficient algorithm to solve the formulated problem and determine the corresponding solution for each device.
	\item Extensive experiments demonstrate that our proposed FIMI can outperform the existing data augmentation FL methods in terms of the device-side energy utilization and the accuracy of the global model after convergence.
\end{itemize}

The remainder of this paper is organized as follows. Section 2 describes related studies. In Section 3, we illustrate the system model of FIMI. 
The theoretical analysis and the corresponding solution are provided in Section 4. 
The experiment evaluations are presented in Section 5. We finally conclude the paper and discuss the future directions in Section 6.

\begin{figure*}[t]\centering
  \includegraphics[width=0.98\textwidth]{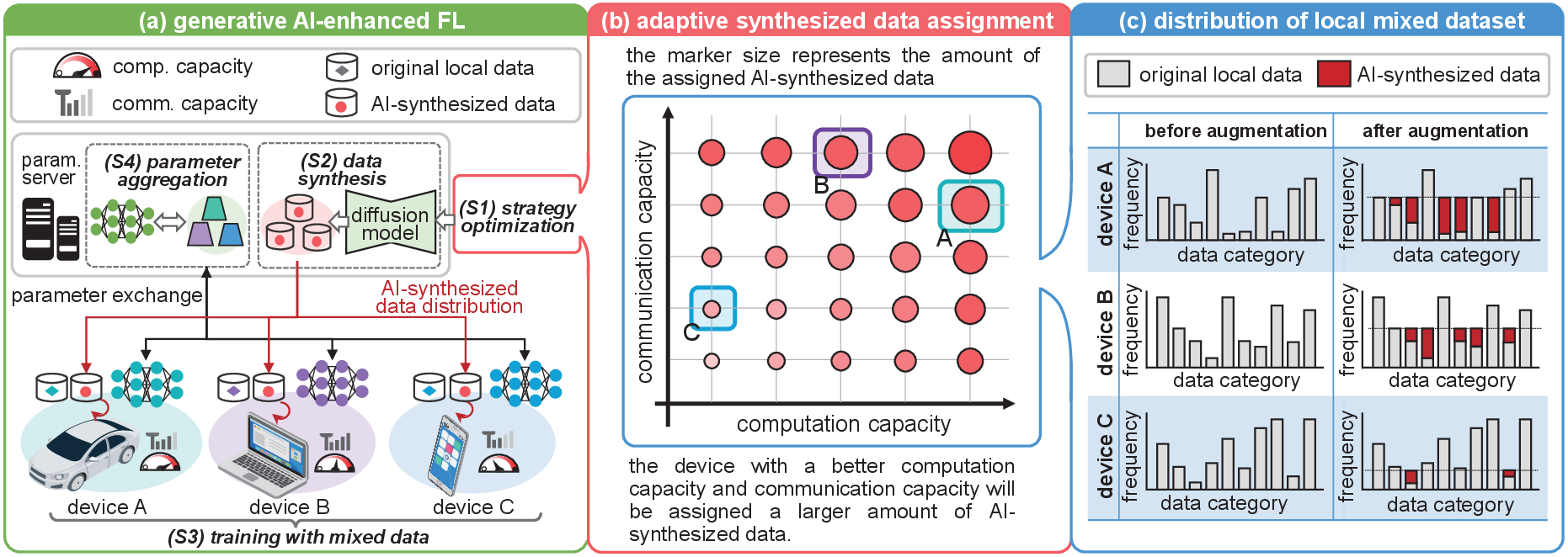}
  \vspace{-5pt}
  \caption{\textbf{left}: FIMI over heterogeneous edge devices. \textbf{middle}: the amounts of synthesized data are personalized for different devices to match their locally available resources. \textbf{right}: each device optimizes its category-wise synthesized data amount to mitigate the locally non-IID feature of data.}\label{fig:sys-model}
\end{figure*}

\section{Related Work}
$\;\;$\emph{1) Generative AI over edge networks}.  Generative AI on edge networks has attracted growing interest for its potential to provide low-latency AI generative content (AIGC) services \cite{xu2023unleashing}. Some studies focus on achieving efficient AIGC service by optimizing resource utilization \cite{10172151} or by improving the freshness of the age of information \cite{xu2023joint}. In addition, diffusion model-integrated reinforcement learning presents a novel approach for solving complex optimization problems \cite{du2023beyond}. Recently, the study in \cite{huang2023federated} incorporates the FL framework to enhance the training and inference process of the AIGC model. However, the potential benefits of generative AI for FL training remain unexplored.

\emph{2) Data augmentation methods for FL}. Data augmentation methods provide an effective approach to address the issue of non-IID data distribution in FL. 
The early study in \cite{zhao2018federated} shows that exchanging a small proportion of local data can alleviate the non-IID issue. However, this approach may raise privacy concerns. In addition, semi-supervised FL methods utilize the unlabelled data to enhance learning performance \cite{itahara2021distillation, wang2023towards}. Recently, generative adversarial networks (GAN) have emerged as a privacy-friendly technique for data compensation in FL \cite{xin2020private, jeong2019multihop}. However, the GAN-based methods may fail to produce photo-realistic data, and most of the existing methods overlook the extra training cost induced by the augmented dataset.

\emph{3) Efficient training methods for FL}. 
Efficient training methods for FL have been widely studied in recent years. One popular approach is the gradient compression, which reduces the amount of data by compressing the gradients \cite{li2023snowball, xu2022adaptive}. Another approach is topology-aware methods, such as hierarchical aggregation \cite{liu2020client, luo2020hfel} and device-to-device communication \cite{hosseinalipour2022multi}, which consider the network topology to optimize the communication pattern and reduce the communication overhead. Furthermore, energy-aware methods, such as wireless power transfer \cite{wu2022non} and resource management \cite{yang2020energy, wu2023split,ko2021joint}, offer a green and sustainable framework for FL training. However, most of the existing studies do not consider the heterogeneity of data distribution among different participating devices, which can result in biased model updates.

\section{System Model}
In this section, we first introduce the structure of FIMI, and then present the mathematical modeling and optimization problems for the data augmentation-based FL training.
\subsection{Structure of FIMI}
We consider an FL scenario comprising a set of devices labeled as $\{1,2,\ldots, I\}$. Each device, denoted as $i$, holds an original local dataset ${\cal D}_i^{\text{loc}}$, where $|{\cal D}_i^{\text{loc}}|=D_i^{\text{loc}}$ represents the number of local training samples. Our goal is to leverage a pre-trained generative model to create additional synthesized data ${\cal D}_i^{\text{gen}}$ for each device $i$, with $|{\cal D}_i^{\text{gen}}|=D_i^{\text{gen}}$.
In traditional FL systems, device $i$ solely relies on its local dataset ${\cal D}_i^{\text{loc}}$ for the local model update. Alternatively, the proposed FIMI adopts a mixed local dataset ${\cal D}_i^{\text{mix}} = {\cal D}_i^{\text{loc}} \cup {\cal D}_i^{\text{gen}}$ to accelerate the convergence.

{\color{black} We focus on the FL task for image classification across $C$ categories (or label types). The data synthesis strategy of device $i$ can be represented as $\{d_{i,c}^{\text{gen}}\}_{\forall c}, (c=1,\ldots,C)$. This strategy should meet the constraint $\sum\nolimits_c d_{i,c}^{\text{gen}} = D_{i}^{\text{gen}}$, where $d_{i,c}^{\text{gen}}$ denotes the amount of synthesized data for the $c$-th category on device $i$. Before the beginning of FL training, a collaborative optimization is undertaken between the server and edge devices to obtain the data synthesis strategy. Following this, the server will employ the corresponding label as input for the diffusion model to generate the required data. After the synthesized data is distributed to each device, the FL training with the mixed dataset is initiated. As shown in Fig.~\ref{fig:sys-model}, the FIMI framework operates as follows.}
\begin{itemize}
    \item \emph{(S1) Strategy optimization}. The server conducts a centralized optimization, as shown in the middle subplot of Fig.~\ref{fig:sys-model}, to determine the synthesized data amount for each device. Each device then calculates the \emph{category-wise} synthesized data amount $\{d_{i,c}^{\text{gen}}\}_{\forall c}, (c=1,\ldots,C)$, as shown in the right subplot of Fig.~{\ref{fig:sys-model}}, to rebalance the data distribution and mitigate the non-IID feature of local data.
    \item \emph{(S2) Data synthesis}. Each device sends its data synthesis request, specifying the category-wise amount of synthesized data, to the server. The server infers the generative model and distributes the required synthesized data to the corresponding devices in parallel.
    \item \emph{(S3) Training with mixed data}. The server distributes the global model to each device. Then, the local device performs the local training with the mixed local dataset, and then uploads its model update back to the server.
    \item \emph{(S4) Parameter aggregation}. The server collects and aggregates the local updates from all devices to renew the global model. The training process jumps to (S3) until the predetermined maximum global iterations are reached.
\end{itemize}
{\color{black}Since the servers are usually equipped with sufficient computational resources and resilient power supply, we do not account for the latency and energy consumption during the data synthesis process. Empirical computation and communication overheads for data synthesis and distribution will be illustrated in Section 5.1.}

\subsection{Learning Performance Analysis}
In this subsection, we quantitatively analyze the relationship between the number of local samples and the overall training performance. We first establish the mathematical formulation that links the amount of local data to the individual learning performance. We then determine the unknown parameters in the formulation through the empirical fitting. Finally, we investigate how the isolated local data from multiple devices can influence the global learning performance. 

\subsubsection{Local Data vs. Local Model Performance} 
Let $\delta _i$ denote the local learning error of device $i$ when training on mixed data ${\cal D}_i^{\text{mix}}$.  According to the previous studies in \cite{chen2018my, wang2020machine, hu2021nothing}, the lower bound of the achievable local error can be modeled as the power-law function with respect to the training sample amount of $|{\cal D}_i^{\text{mix}}|$. The relationship between $D_i^{\text{loc}} + D_i^{\text{gen}}$ and $\delta _i$ can be expressed as
\begin{equation}
\label{eq:local_error}
    \delta _i = \alpha (D_i^{\text{loc}} + D_i^{\text{gen}})^{ - \beta } - \gamma,
\end{equation}
where $\alpha, \beta$, and $\gamma$ are three positive hyper-parameters that can be determined through experiments.

\subsubsection{Parameter Fitting on Proxy Task} 
\begin{figure}[t]\centering
  \includegraphics[width=0.48\textwidth]{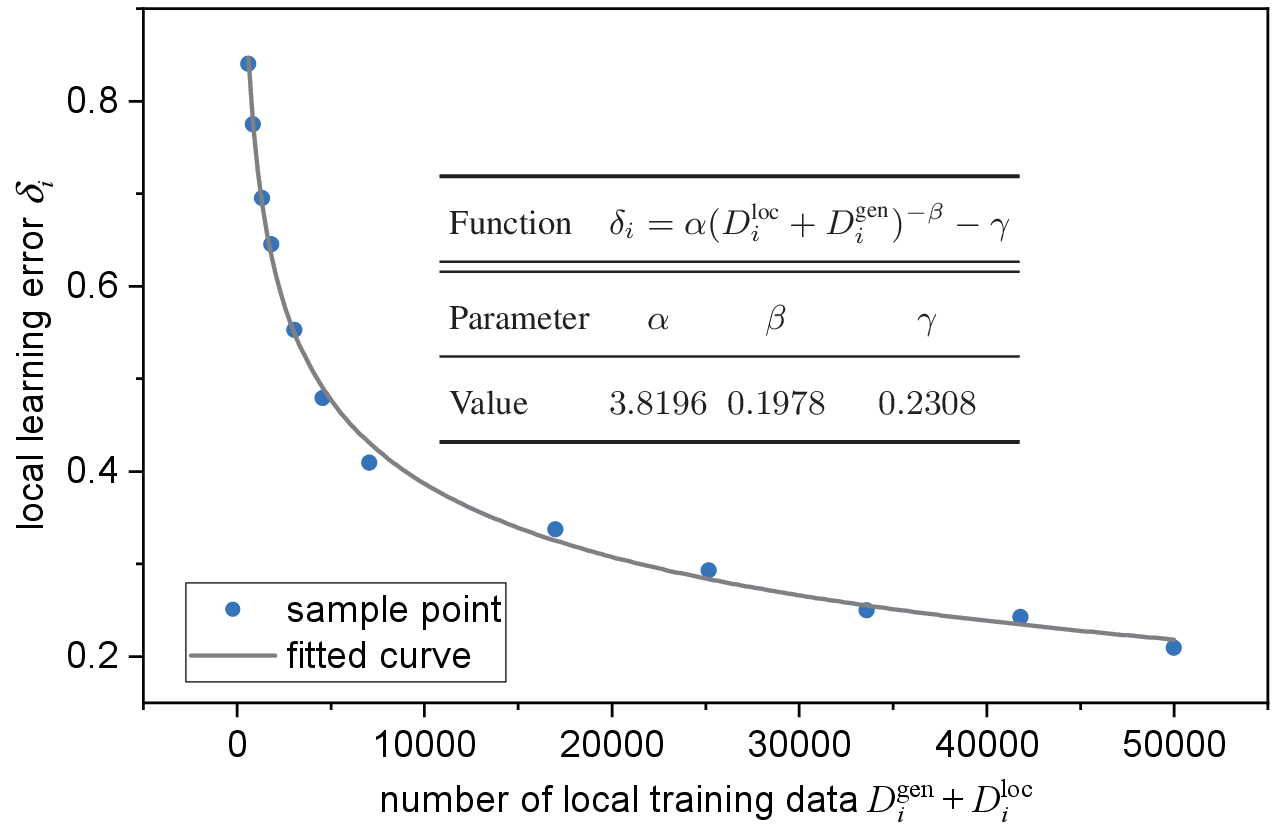}
\caption{Local learning error $\delta _i$ with respect to local training data amount.} \label{fig:fit-curve}
\end{figure}

We next focus on determining the values of $\{\alpha, \beta, \gamma\}$. However, since the local dataset is not visible to the server, direct parameter fitting on the target learning task by the server can raise privacy concerns. We propose an alternative approach by determining these parameters using a public dataset. Specifically, we employ the classification task on the CINIC10 dataset as the \emph{proxy} task for parameter fitting \cite{darlow2018cinic}. By assessing the learning error of well-trained models on a variety of data amounts, we obtain the numerical results as shown in Fig.~\ref{fig:fit-curve}. {\color{black}It is noticed that this \emph{one-time} parameter fitting process can be conducted offline by the server, and the results of fitting can be generalized to \emph{many} learning tasks with different datasets.}

\subsubsection{Global Data vs. Global Model Performance} After estimating the local learning performance $\{\delta _i\}_{\forall i}$, the average local training error $\overline{\delta}$ can be calculated as
\begin{equation}\label{eq:avg_loc_error}
    \overline{\delta} = {\frac{1}{I}}\sum\limits_{i = 1}^I {\delta _i}.
\end{equation}
Based on the theoretical analysis in \cite{ma2017distributed, tran2019federated}, when training with the average local training error $\overline{\delta}$, the number of global iterations $N$ to reach a pre-determined global learning error $\Delta$ is bounded by 
\begin{equation}
    N = \frac{{\zeta \ln (1/\Delta )}}{{1 - \overline{\delta} }},
\end{equation}
where $\zeta$ is a positive constant parameter. Furthermore, the relationship between the average local learning error $\overline{\delta}$ and the global learning error $\Delta$ can be captured by
\begin{equation}\label{eq:loc-glb-error}
    \Delta = \exp \Big(\frac{{N(\overline{\delta}  - 1)}}{\zeta }\Big).
\end{equation}

By combing Eqns.~(\ref{eq:local_error}), (\ref{eq:avg_loc_error}) and (\ref{eq:loc-glb-error}), we bridge the relationship between the amount of local mixed dataset and the global learning performance.
\subsection{FIMI over Wireless Networks}

\subsubsection{Computation Model} Let $\omega$ denote the computation workload required for a single training sample and $\tau$ represent the local epoch. According to \cite{tran2019federated, li2023anycostfl, yang2020energy}, the computation-related configurations of device $i$ can be profiled by two parameters, including the computing frequency $f_i$ and the hardware energy coefficient $\epsilon_i$. Thus, the single-round energy consumption for the local model training at device $i$ can be estimated by
\begin{equation}\label{eq:local_cmp_energy}
    E_i^{\text{cmp}} = \tau \epsilon_i{\omega}(D_i^{\text{loc}} + D_i^{\text{gen}}) f^2_i.
\end{equation}
Furthermore, the required latency for a single-round local model training of device $i$ is
\begin{equation}\label{eq:loc-cmp-time}
    T^{\text{cmp}}_i = \frac{\tau{\omega}(D_i^{\text{loc}} + D_i^{\text{gen}})}{f_i}.
\end{equation}

\subsubsection{Communication Model} 
We next focus on the device-side uplink parameter transmission. We consider the frequency division multiple access (FDMA) scheme with a preset total bandwidth of $B$. For the device $i$ allocated with a sub-bandwidth of $b_i$, its achievable transmitting rate for sending the local update to the server can be given by 
\begin{equation}\label{eq:com-rate}
    r_i = b_i \log_2\big(1+\frac{g_iP_i}{N_0b_i}\big),
\end{equation}
where $g_i$ is the device-to-server channel gain, $P_i$ is the transmitting power, and $N_0$ is the power spectral density of background Gaussian noise. Here, we utilize $S$ to denote the data size of the model update. 
The single-round latency required for device $i$ to send its model update to the server is estimated by
\begin{equation}\label{eq:com-local-time}
    T^{\text{com}}_i = \frac{S}{r_i}.
\end{equation}
Moreover, the single-round energy consumption for device $i$ to send its model update to the server can be calculated by 
\begin{equation}
    E^{\text{com}}_i = \frac{SP_i}{r_i}.
\end{equation}

\subsubsection{Problem Formulation} Based on the above modeling, the energy consumption of all the devices for one-round training can be expressed as
\begin{equation}
    E^{\text{round}} = \sum\limits_{i = 1}^I E^{\text{cmp}}_i + \sum\limits_{i = 1}^I E^{\text{com}}_i.
\end{equation}
Meanwhile, the system-wise learning latency for one-round training is calculated by{\color{black}
\begin{equation}
    T^{\text{round}} = \max \limits_{i}\big\{T^{\text{cmp}}_i + T^{\text{com}}_i\big\}.
\end{equation}}
To optimize the proposed FIMI, we investigate an energy minimization problem with a pre-determined learning performance constraint. Specifically, given multiple devices with heterogeneous resources in terms of computation, communication, and data distribution, our objective is to optimize the optimal data synthesis strategy (i.e., $\{D_{i}^{\text{gen}}\}, i=1,\ldots, I$) and resource utilization policy (i.e., $\{b_i, f_i, P_i \}, i=1,\ldots, I$) for each device. Mathematically, we aim at studying the following problem for each global round.
\begin{subequations}\label{eqn:pbm-1}
\begin{align}
    ({\text{P1}}) && &~~~~~\min ~~  E^{\text{round}}    &\tag{\ref{eqn:pbm-1}}\\
    {\text{subject to:}} && &~~~~~~\Delta \le \Delta^{\max} ,&\label{eqn:p1-ctr-1}\\
    {} && &~~~~  T^{\text{round}} \le T^{\max},\label{eqn:p1-ctr-2}\\
    {}&& &~~0 \le {D _{i}^{\text{gen}}} \le D^{\max},\forall i, & \label{eqn:p1-ctr-4}\\
    {}&& &~~~0 \le {f _{i}}  \le f_i^{\max},\forall i, & \label{eqn:p1-ctr-5}\\
    {}&& &~~~~ \sum\nolimits_{i=1}^I  b_i \le B, &\label{eqn:p1-ctr-3}\\
    {}&& &~~~0 \le P_{i} \le P^{\max}_{i},\forall i, &\label{eqn:p1-ctr-6}\\
	{{\textrm{variables:}}}&& &~~~~\{D_{i}^{\text{gen}}, f_i, b_i, P_i \}_{\forall i},&\nonumber
\end{align}
\end{subequations}
where $\Delta^{\max}, T^{\max}$, and $D^{\max}$ are the pre-determined maximum allowable global error, the maximum latency for single round learning, and the maximum amount of synthesized data for each device, respectively. Note that the data synthesis procedure is conducted by the server with powerful computing capabilities. Therefore, we solely focus on the device-side energy consumption in Problem (P1).

\subsection{Analysis and Decomposition}
Before proposing the algorithms to solve Problem (P1) in Section 4, we present the analysis and decomposition in this subsection. We first leverage two lemmas to simplify the main problem. After that, we decompose the simplified problem into three sub-problems.

To facilitate the analysis, we present the following lemmas.
\begin{lem}\label{lem:1}
The equality in Constraint (\ref{eqn:p1-ctr-1}) always holds when the optimal data synthesis strategy $\{D_{i}^{\text{gen},\ast}\}_{\forall i}$ is achieved.
\end{lem}
\vspace{-2pt}
\begin{proof*}
The lemma can be proved by showing contradiction. 
Given the optimal data synthesis strategy $\{D_{i}^{\text{gen},\ast}\}_{\forall i}$, suppose that the achieved global learning error is less than the maximum allowable error, i.e., $\Delta(\{D_{i}^{\text{gen},\ast}\}_{\forall i}) < \Delta ^{\max}$. Based on Eqns.~(\ref{eq:local_error}) and (\ref{eq:loc-glb-error}), the global learning error $\Delta$ decreases with $D_{i}^{\text{gen}}$. Therefore, a feasible solution can always be constructed by reducing the synthesized data of device $i_0$ from $D_{i_0}^{\text{gen},\ast}$ to $D_{i_0}^{\text{gen},\prime}$, such that $\Delta(\{D_{i}^{\text{gen},\ast}\}_{\forall i, i\neq i_0}\cup D_{i_0}^{\text{gen},\prime}) = \Delta ^{\max}$. As $E^{\text{round}}$ increases with $D_i^{\text{gen}}$, the energy consumption of the new solution $\{D_{i}^{\text{gen},\ast}\}_{\forall i, i\neq i_0}\cup D_{i_0}^{\text{gen},\prime}$ will be lower than that of the optimal one. This completes the proof.
\end{proof*}
\begin{lem}
The equality in Constraint (\ref{eqn:p1-ctr-2}) always holds under the optimal resource utilization policy $\{b_i^\ast, f_i^\ast, P_i^\ast \}_{\forall i}$.
\end{lem}
\vspace{-2pt}
\begin{proof*}
The lemma can be proved by showing contradiction. Suppose that under the optimal resource utilization policy $\{b_i^\ast, f_i^\ast, P_i^\ast \}_{\forall i}$, the single-round latency required is smaller than the maximum latency, i.e., $T^{\text{round}} < T^{\max}$. According to Eqn.~(\ref{eq:loc-cmp-time}), the training latency increases with $f_i$. Therefore, a feasible solution can always be constructed by reducing the computing frequency of device $i_0$ from $f_{i_0}^\ast$ to $f_{i_0}^\prime$. Similar to the proof of Lemma \ref{lem:1}, the new solution will consume less energy than the optimal one. This completes the proof.
\end{proof*}

By combing Eqn.~(\ref{eq:loc-glb-error}) and the above two lemmas, Problem (P1) can be simplified as
\begin{subequations}\label{eqn:pbm-1-1}
\begin{align}
    ({\text{P2}}) && &~~~~~~~\min ~~  E^{\text{round}}    &\tag{\ref{eqn:pbm-1-1}}\\
    {\text{subject to:}} && &\sum\limits_{i = 1}^I  \delta_i = I + \frac{\zeta I}{N}\ln \big({\Delta^{\max}}\big),&\label{eqn:p1-ctr-1-1}\\
    {} && &~~~T^{\text{cmp}}_i + T^{\text{com}}_i = T^{\max},\label{eqn:p1-ctr-2-1}\\
    {}&& &~~ \text{Constraints (\ref{eqn:p1-ctr-4})-(\ref{eqn:p1-ctr-6})},\\
	{{\textrm{variables:}}}&& &~~~~~\{D_{i}^{\text{gen}}, f_i, b_i, P_i \}_{\forall i}.&\nonumber
\end{align}
\end{subequations} 

Moreover, by determining the latency for both local model training and local update transmission for each device (i.e., $\{T^{\text{cmp}}_i, T^{\text{com}}_i\}_{\forall i}$), Problem (P2) can be decomposed into two distinct sub-problems as described in the following. 

The first sub-problem, denoted as Problem (P3), aims at minimizing all devices' energy consumption for their local model training, which is expressed as follows:
\begin{subequations}\label{eqn:pbm-cmp}
\begin{align}
    ({\text{P3}}) && &~~~\min \limits_{\{D_{i}^{\text{gen}}, f_i\}_{\forall i}} \sum\limits_{i = 1}^I \tau\epsilon_i{\omega}(D_i^{\text{loc}} + D_i^{\text{gen}}) f^2_i    &\tag{\ref{eqn:pbm-cmp}}\\
    {\text{subject to:}} && &~~~ \frac{\tau{\omega}(D_i^{\text{loc}} + D_i^{\text{gen}})}{f_i} = T^{\text{cmp}}_i, \forall i ,\label{eqn:p2-ctr-2}\\
    {}&& &\text{Constraints (\ref{eqn:p1-ctr-4}), (\ref{eqn:p1-ctr-5}), and (\ref{eqn:p1-ctr-1-1})}.
\end{align}
\end{subequations}

The second sub-problem, denoted as Problem (P4), aims at minimizing all devices' energy consumption for uploading their locally trained models as follows:
\begin{subequations}\label{eqn:pbm-4}
\begin{align}
    ({\text{P4}}) && &~~~~~~\min \limits_{\{ b_i, P_i \}_{\forall i}} ~  \sum\limits_{i = 1}^I P_i T_i^{\text{com}}    &\tag{\ref{eqn:pbm-4}}\\
    {\text{subject to:}} && & \frac{S}{{b_i \log_2\big(1+\frac{g_iP_i}{N_0b_i}\big)}} = T_i^{\text{com}}, \forall i, \label{eqn:p4-ctr-2}\\
    {}&& & ~~\text{Constraints (\ref{eqn:p1-ctr-3}) and (\ref{eqn:p1-ctr-6})}.
\end{align}
\end{subequations}

After solving the above two sub-problems, we can obtain $\{D_{i}^{\text{gen}}, f_i, b_i, P_i \}_{\forall i}$ with given $\{T_i^{\text{cmp}}, T_i^{\text{com}}\}$. Then, we continue to determine the optimal $\{T_i^{\text{cmp}, \ast}, T_i^{\text{com}, \ast}\}_{\forall i}$ for each device.
Instead of directly optimizing $\{T_i^{\text{cmp}}, T_i^{\text{com}}\}$, we introduce an auxiliary variable for each device to reduce the dimension of optimization variables. Let $\eta _i \in (0, 1)$ denote the time splitting factor for device $i$. Then, we have $T_i^{\text{cmp}} = \eta_iT^{\max}$ and $T_i^{\text{com}} = (1-\eta_i)T^{\max}$. Thus, our goal can be transformed into a top-layer searching problem that optimizes the splitting factor $\{ \eta_i \}_{\forall i}$ for each device as follows:
\begin{subequations}\label{eqn:pbm-6}
\begin{align}
    ({\text{P5}}) && &~~~~~~\min ~  E^{\text{round}}\big(\{\eta_i\}_{\forall i}\big)    &\tag{\ref{eqn:pbm-6}}\\
    {\text{subject to:}} && & ~~~~~\eta_i^{\min} \le \eta_i \le \eta_i^{\max},\forall i, &\label{eqn:p6-ctr-6}\\
	{{\textrm{variables:}}}&& &~~~~~~~~~~~~~\{ \eta_i \}_{\forall i},&\nonumber
\end{align}
\end{subequations}
where $\eta_i^{\min}$ and $\eta_i^{\max}$ are the lower and the upper bounds of the time splitting factor, respectively. Specifically, by combing Eqns.~(\ref{eq:loc-cmp-time}), (\ref{eq:com-rate}) and (\ref{eq:com-local-time}), the value of $\eta_i^{\min}$ can be obtained by
\begin{equation}
    \eta _i ^{\min} =  \frac{{\tau\omega {D_i^{\text{loc}}}}}{{T^{\max }f_i^{\max}}},
\end{equation}
and $\eta_i^{\max}$ can be calculated by
\begin{equation}
    \eta _i ^{\max} = 1 - \frac{S}{{T^{\max }B{{\log }_2}(1 + \frac{{{g_i}P_i^{\max }}}{{{N_0}B}})}}.
\end{equation}
After decomposing the main problem into two sub-problems, we will proceed to present the solution for each problem in Section 4.

\section{Proposed Algorithm for FIMI}
\subsection{Solution for Problem (P3)}
The procedure for solving Problem (P3) is outlined as follows. We first simplify the problem by variable substitutions. After that, we leverage the convex optimization technique to obtain the optimal solution.

Based on Eqn.~(\ref{eq:local_error}), $D_i^{\text{gen}}$ can be expressed as a function of $\delta _i$ as follows:
\begin{equation}\label{eq:var-subs-1}
    D_i^{\text{gen}}  = \Big(\frac{\gamma + \delta _i}{\alpha}\Big)^{-\frac{1}{\beta}} - D_i^{\text{loc}}.
\end{equation} 
Moreover, by combing Constraint (\ref{eqn:p2-ctr-2}) and Eqn.~(\ref{eq:var-subs-1}), $f_i$ can be further represented by $\delta _i$ as
\begin{equation}\label{eq:var-subs-2}
    {f_i} = \frac{{\tau \omega {{(\frac{{\gamma  + {\delta _i}}}{\alpha })}^{ - \frac{1}{\beta }}}}}{{T_i^{{\rm{cmp}}}}}.
\end{equation}

Next, by substituting Eqns.~(\ref{eq:var-subs-1}) and (\ref{eq:var-subs-2}) into Problem (P3), the problem is simplified as follows:
\begin{subequations}\label{eqn:pbm-cmp-p3}
\begin{align}
    ({\text{P6}}) && &~~~\min \limits_{\{\delta _i\}_{\forall i}} \sum\limits_{i = 1}^I  \rho_i (\gamma + \delta_i)^{-\frac{3}{\beta}}    &\tag{\ref{eqn:pbm-cmp-p3}}\\
    {\text{subject to:}} && &\sum\limits_{i = 1}^I  \delta_i = I + \frac{\zeta I}{N}\ln \big({\Delta^{\max}}\big) ,&\label{eqn:p3-ctr-1}\\
    {}&& &~~~~~{\delta _i ^{\min}} \le {\delta _i} \le {\delta _i^{\max}},\forall i, & \label{eqn:p3-ctr-4}
\end{align}
\end{subequations}
where $\rho_i$ is an intermediate constant calculated by
\begin{equation}
    \rho_i  = \frac{\epsilon_i {(\tau\omega)^3}}{\big(T^{\text{cmp}}_i\big)^2 \alpha^{-\frac{3}{\beta}}} > 0. 
\end{equation}
Here, the lower bound of $\delta _i$ can be obtained by
\begin{equation}
    \delta _i^{\min} =  \alpha \Big(\min \Big\{\frac{f_i^{\max} T_i^{\text{cmp}}}{\tau\omega} , D_i^{\text{loc}} + D^{\max}\Big\}\Big)^{ - \beta } - \gamma, 
\end{equation}
and the upper bound of $\delta _i$ is calculated as
\begin{equation}
    \delta _i^{\max} =  \alpha (D_i^{\text{loc}})^{ - \beta } - \gamma.
\end{equation}

Note that the solution to Problem (P6) relies on Constraint (\ref{eqn:p3-ctr-1}). If $\sum \nolimits_{i=1}^I\delta_i^{\min} > I + \frac{\zeta I}{N}\ln \big({\Delta^{\max}}\big)$ or $\sum \nolimits_{i=1}^I\delta_i^{\max} < I + \frac{\zeta I}{N}\ln \big({\Delta^{\max}}\big)$, then Problem (P6) is infeasible. In the following, we focus on the practical case that $\sum \nolimits_{i=1}^I\delta_i^{\min} \le I + \frac{\zeta I}{N}\ln \big({\Delta^{\max}}\big) \le \sum \nolimits_{i=1}^I\delta_i^{\max}$.

\begin{thm}[Solution for Problem (P6)]
Suppose that a feasible scenario where inequality $\sum \nolimits_{i=1}^I\delta_i^{\min} \le I + \frac{\zeta I}{N}\ln \big({\Delta^{\max}}\big) \le \sum \nolimits_{i=1}^I\delta_i^{\max}$ holds. The optimal solution to Problem (P6) can be expressed as \footnote{The operation of $[x]_{a}^{b}=\min\{b,\max\{x,a\}\}$.}
\begin{align}
\begin{split}
    {\delta _{i}^\ast} =  \bigg[ {\Big( {\frac{{3{\rho _i}}}{{\beta \nu ^\ast }}} \Big)^{\frac{\beta }{{\beta  + 3}}}}\bigg]^{\delta _{i}^{\max }}_{\delta _{i}^{\min }} , \;\forall i,
\end{split}
\label{eq:thm1}
\end{align}
where the variable $\nu ^\ast$ is determined through a search process in order to find the optimal solution $\{\delta_i^\ast\}_{\forall i}$ that satisfies Constraint (\ref{eqn:p3-ctr-1}).
\label{thm:1}
\end{thm}
\begin{proof*}
It can be verified that Problem (P6) is a convex optimization problem with respect to $\delta _i$. We next employ Karush–Kuhn–Tucker (KKT) the necessary condition for achieving the optimality. By introducing $\{\lambda _i\}_{\forall i}$ and $\{\psi _i\}_{\forall i}$ as the Lagrange multipliers for inequality constraints, and $\nu$ as the multiplier for the equality constraint, we have
\begin{subequations}
\begin{empheq}[left=\empheqlbrace]{align}
&\lambda _i \ge 0, \; \lambda _i (\delta_i^{\min} - \delta _i)=0, \forall i, \label{eqn:p4-kkt-1}\\
&\psi _i \ge 0, \; \psi _i(\delta _{i} - \delta_{i}^{\max})=0, \forall i,\label{eqn:p4-kkt-2}\\
&-\frac{3{\rho _i}}{{\beta} (\delta _i+{\gamma})^\frac{{\beta}+3}{{\beta}}} - {\lambda _i} + {\psi _i}  + \nu = 0, \forall i,\label{eqn:p4-kkt-3}\\
&\text{Constraints (\ref{eqn:p3-ctr-1}) and (\ref{eqn:p3-ctr-4})}.
\end{empheq}
\end{subequations}

By putting Eqn.~(\ref{eqn:p4-kkt-3}) into Eqns.~(\ref{eqn:p4-kkt-1}) and (\ref{eqn:p4-kkt-3}), we have
\begin{subequations}
\small
\begin{empheq}[left=\empheqlbrace]{align}
&\lambda _i \ge 0, \; \psi_i \ge 0, \forall i, \label{eqn:p4-kkt-s1}\\
&\bigg(\underbrace{ -\frac{3{\rho _i}}{{\beta} (\delta _i+{\gamma})^\frac{{\beta}+3}{{\beta}}} + {\psi _i}  + \nu}_{A_1} \bigg)(\delta_i^{\min} - \delta _i) = 0, \forall i, \label{eqn:p4-kkt-s2}\\
&\bigg( \underbrace{\frac{3{\rho _i}}{{\beta} (\delta _i+{\gamma})^\frac{{\beta}+3}{{\beta}}} + {\lambda _i} - \nu}_{A_2} \bigg)\left( \delta _{i} - \delta_{i}^{\max} \right) = 0, \forall i. \label{eqn:p4-kkt-s3}
\end{empheq}
\end{subequations}

Furthermore, the solution to Problem (P6) can be categorized into the following three cases, depending on the value of $\nu$. 
\begin{itemize}
    \item When $\nu > \frac{3{\rho _i}}{{\beta} (\delta _i^{\min}+{\gamma})^\frac{{\beta}+3}{{\beta}}}$, we have $A_1>0$. Based on Eqn.~(\ref{eqn:p4-kkt-s1}), we obtain $\delta _{i}=\delta _{i}^{\min}$.
    \item When $\nu < \frac{3{\rho _i}}{{\beta} (\delta _i^{\max}+{\gamma})^\frac{{\beta}+3}{{\beta}}}$, we have $A_2>0$. Thus, we get $\delta _{i}=\delta _{i}^{\max}$.
    \item When $  \frac{3{\rho _i}}{{\beta} (\delta _i^{\max}+{\gamma})^\frac{{\beta}+3}{{\beta}}} \le \nu \le \frac{3{\rho _i}}{{\beta} (\delta _i^{\min}+{\gamma})^\frac{{\beta}+3}{{\beta}}}$, we have $\lambda _i=0$ and $ \psi _i=0$. Thus $\nu = \frac{3{\rho _i}}{{\beta} (\delta _i+{\gamma})^\frac{{\beta}+3}{{\beta}}}$.
\end{itemize}
By summarizing the above three cases into one equation, we obtain the solution in Eqn.~(\ref{eq:thm1}). Thus, we complete the proof.
\end{proof*}
We next aim to find the value of $\nu^\ast$ that satisfies Constraint (\ref{eqn:p3-ctr-1}) according to Theorem \ref{thm:1}. Based on Eqn.~(\ref{eq:thm1}), the optimal value $\nu ^ \ast$ can be expressed as a function of $\delta _i^ \ast$ as follows:
\begin{equation}
    {\nu ^ \ast } = \frac{{3{\rho _i}}}{\beta }{\big( {\delta _i^ \ast } \big)^{ - \frac{{\beta  + 3}}{\beta }}}.
\end{equation}
Note that $\nu ^ \ast$ decreases as $\delta _i^ \ast$ increases for $\delta _i^ \ast > 0$. Thus, the search range $[\nu ^{\min}, \nu ^{\max}]$ for $\nu ^ \ast$ can be given by
\begin{subequations}\label{eq:nu-range}
\begin{empheq}[left=\empheqlbrace]{align}
& {\nu ^ {\min} } =\min \limits_i {\Big\{ \frac{{3{\rho _i}}}{\beta }{\big( {\delta _i^ {\max} } \big)^{ - \frac{{\beta  + 3}}{\beta }}}  \Big\}}, \label{eq:nu-range-1}\\
& {\nu ^ {\max} } =\max \limits_i {\Big\{ \frac{{3{\rho _i}}}{\beta }{\big( {\delta _i^ {\min} } \big)^{ - \frac{{\beta  + 3}}{\beta }}}  \Big\}}.\label{eq:nu-range-2}
\end{empheq}
\end{subequations}
According to Eqn.~(\ref{eq:thm1}),  the value of $\sum \nolimits_{i=1}^{I} \delta_i$ is non-increasing with respect to $\nu$. Therefore, the value of $\nu ^\ast$ can be efficiently obtained by the bi-section search. Furthermore, the algorithm for solving Problem (P3) is shown in Algorithm \ref{alg:bin-search}. Specifically, the computational complexity for Algorithm \ref{alg:bin-search} can be approximated by ${\cal O}\big(\log_2(\nu^{\max} - \nu^{\min})\big)$.
\begin{algorithm}
\SetAlgoLined
\SetKwInput{KwInput}{Input}
\SetKwInput{KwOutput}{Output}
\SetKwRepeat{KwRepeat}{repeat}{until}
\KwInput{$\nu ^ {\min}, \nu ^ {\max}$, and a small positive number $\varepsilon$.}
\KwRepeat{$|\nu^{\max} - \nu^{\min}| \leq \varepsilon$}{
    Update $\nu = (\nu^{\max} + \nu^{\min})/2$\;
    Compute $\{\delta _i\}_{\forall i}$ based on Eqn.~(\ref{eq:thm1})\;
    \textbf{if} $\sum \nolimits_{i=1}^{I} \delta_i > I + \frac{\zeta I}{N}\ln \big({\Delta^{\max}}\big)$ \textbf{then} update $\nu^{\min} = \nu$ \textbf{else} update $\nu^{\max} = \nu$\;
}
Set $\nu^\ast = \nu$, and acquire $\{\delta_i^\ast\}_{\forall i}$ based on Eqn.~(\ref{eq:thm1})\;
Compute $\{D^{\text{gen},\ast}_i\}_{\forall i}$ based on Eqn.~(\ref{eq:var-subs-1}), and obtain $\{f_i^{\ast}\}_{\forall i}$ based on Eqn.~(\ref{eq:var-subs-2})\;
\textbf{return} $\{D^{\text{gen},\ast}_i, f_i^{\ast}\}_{\forall i}$\\
\caption{Bisection search for Problem (P3)}
\label{alg:bin-search}
\end{algorithm}

\subsection{Solution for Problem (P4)}
To solve Problem (P4), we begin by transforming the problem into a simplified form that optimizes the bandwidth allocation. 

Based on Constraint~(\ref{eqn:p4-ctr-2}), $P_i$ can be represented by $b_i$ as
\begin{equation}
\label{eq:trans_power}
     P_i = \frac{N_0b_i}{g_i}\big(2^{\frac{S}{b_i T_i^{\text{com}}}} - 1\big).
\end{equation}
Note that $P_i$ is decreasing with respect to $b_i$. Thus, by incorporating Constraint (\ref{eqn:p1-ctr-6}), the lower bound of $b_i$ can be expressed as
\begin{equation}
\label{eq:lower_bandw}
    b_i \ge b_i^{\min} = - \frac{S\ln2}{T_i^{\text{com}} W \Big( -\frac{\kappa _i}{T^{\text{com}}_i}\exp\big(-\frac{\kappa _i}{T^{\text{com}}_i}\big) \Big) + \kappa_i } ,
\end{equation}
where
$\kappa _i = \frac{N_0 S \ln2}{g_iP^{\max}_i}$ is an intermediate variable and $W(\cdot)$ is the Lambert function.

By putting Eqn.~(\ref{eq:trans_power}) into Eqn.~(\ref{eqn:pbm-4}) and replacing Constraint (\ref{eqn:p1-ctr-6}) to Eqn.~(\ref{eq:lower_bandw}), Problem (P4) can be simplified as follows:
\begin{subequations}\label{eqn:pbm-5}
\begin{align}
    ({\text{P7}}) && &~\min \limits_{\{ b_i\}_{\forall i}} ~  \sum\limits_{i = 1}^I \frac{N_0b_i T_i^{\text{com}}}{g_i}\big(2^{\frac{S}{b_i T_i^{\text{com}}}} - 1\big)        &\tag{\ref{eqn:pbm-5}}\\
    {\text{subject to:}} && & ~~~~~~~~~~ \sum\nolimits_{i=1}^I  b_i = B, &\label{eqn:p5-ctr-3}\\
    {}&& &~~~~~~~~~~~ b_{i} \ge b^{\min}_{i},\forall i, &\label{eqn:p5-ctr-6}.
\end{align}
\end{subequations}

\begin{thm}[Solution for Problem (P7)]
The optimal solution to Problem (P7) can be expressed as
\begin{align}
\begin{split}
    {b _{i}^\ast} =  \max\big\{ b_i^{\min}, b_i(\varpi ^\ast)\big\}, \;\forall i,
\end{split}
\label{eq:thm2}
\end{align}
where the value of $\varpi ^\ast$ is determined through a search process, and the value of $b_i(\varpi ^\ast)$ is obtained by solving the equation $ {Q(b_i) + \varpi^\ast =0}$. Here, the function $Q(b_i)$ is defined as
\begin{align}\label{eq:func-varpi}
    \begin{split}
        Q(b_i) = \frac{N_0T^{\text{com}}_i\Big(2^\frac{S}{T^{\text{com}}_ib_i}-1\Big)}{g_i}-\frac{\ln(2)N_0S{\cdot}2^\frac{S}{T^{\text{com}}_ib_i}}{g_ib_i}.
    \end{split}
\end{align}
The optimal solution $\{b _{i}^\ast\}_{\forall i}$ under $\varpi ^{\ast}$ should satisfy Constraint (\ref{eqn:p5-ctr-3}).
\label{thm:2}
\end{thm}
\begin{proof*}
It can be verified that Problem (P7) is a convex optimization problem. We further utilize KKT conditions to obtain the necessary condition for achieving the optimality. By introducing $\{\varphi _i\}_{\forall i}$ as Lagrange multipliers for inequality constraints and $\varpi $ as the multiplier for the equality constraint, we have 
\begin{subequations}
\small
\begin{empheq}[left=\empheqlbrace]{align}
&\varphi _i \ge 0, \; \varphi _i (b_i^{\min} - b _i)=0, \forall i, \label{eqn:p5-kkt-1}\\
&\frac{N_0T^{\text{com}}_i\Big(2^\frac{S}{T^{\text{com}}_ib_i}-1\Big)}{g_i}-\frac{\ln(2)N_0S{\cdot}2^\frac{S}{T^{\text{com}}_ib_i}}{g_ib_i} \nonumber\\&\quad\quad\quad\quad\quad\quad\quad\quad\quad\quad\quad\quad- {\varphi _i} + \varpi = 0, \forall i,\label{eqn:p5-kkt-3}\\
&\text{Constraints (\ref{eqn:p5-ctr-3}) and (\ref{eqn:p5-ctr-6})}.    
\end{empheq}
\end{subequations}
By putting Eqn.~(\ref{eqn:p5-kkt-3}) into Eqn.~(\ref{eqn:p5-kkt-1}), we have
\begin{align}\label{eq:com-kkt}
    \begin{split}
\big(Q(b_i) + \varpi\big) \cdot (b_i^{\min} - b _i) = 0,
    \end{split}
\end{align}
where $Q(b_i)$ is a function of $b_i$, and it is expressed as
\begin{equation}
    Q(b_i) = \frac{N_0T^{\text{com}}_i\Big(2^\frac{S}{T^{\text{com}}_ib_i}-1\Big)}{g_i}-\frac{\ln(2)N_0S{\cdot}2^\frac{S}{T^{\text{com}}_ib_i}}{g_ib_i}.
\end{equation}
Let $b_i(\varpi)$ denote the solution of $ {Q(b_i) + \varpi =0}$. Based on Eqn.~(\ref{eq:com-kkt}), the optimal solution to Problem (P6) is given by
\begin{equation}\label{eqn:5-opt}
    {b _{i}^\ast} =  \max\big\{ b_i^{\min}, b_i(\varpi ^\ast)\big\}, \;\forall i.
\end{equation}
Thus, we complete the proof.
\end{proof*}

We next investigate how the solution of ${Q(b_i) + \varpi =0}$ is affected by $\varpi$. Specifically, the first-order derivative of $ {Q(b_i)}$ with respect to $b_i$ is expressed as
\begin{equation}\label{eq:first-derive}
    \frac{{\partial {H }({b_i})}}{{\partial {b_i}}} = \frac{{{{\ln }^2}\left( 2 \right) {S^2}{N_0} \cdot {2^{\frac{S}{{T_i^{{\text{com}}}{b_i}}}}}}}{{T_i^{{\text{com}}}{g_i}b_i^3}} > 0.
\end{equation}
Since $ {Q(b_i)}$ is an increasing function with respect to $b_i$, the solution for ${Q(b_i) + \varpi =0}$, denoted as $b_i(\varpi)$, will decrease with the increase of $\varpi$. Therefore, the value of $b_i(\varpi) \in [b_i^{\min}, B]$ can be obtained through the bisection search. 

After that, our goal turns to search for $\varpi^\ast$ that satisfies Constraint (\ref{eqn:p5-ctr-3}). Since $\varpi$ decreases with $b_i(\varpi)$, the search range $[\varpi^{\min}, \varpi^{\max}]$ for $\varpi$ can be calculated by
\begin{equation}\label{eq:range-varpi}
   \varpi^{\min} = Q(B), ~\text{and } \varpi^{\max} = \max\limits_i \big\{Q(b_i^{\min})\big\}.
\end{equation}
By combing Eqn.~(\ref{eq:thm2}) and the monotonicity of $b_i(\varpi)$, the value of $\sum\nolimits_{i=1}^I  b_i$ is non-increasing with respect to $\varpi$. Thus, the value of $\varpi ^{\ast}$ can be obtained by an outer bisection search.

\begin{algorithm}
\SetAlgoLined
\SetKwInput{KwInput}{Input}
\SetKwInput{KwOutput}{Output}
\SetKwRepeat{KwRepeat}{repeat}{until}
\KwInput{$\varpi ^{\min}, \varpi ^{\max}$, and error bound $\varepsilon$.}
\KwRepeat{$|\varpi^{\max} - \varpi^{\min}| \leq \varepsilon$}{
    Update $\varpi = (\varpi^{\max} + \varpi^{\min})/2$\;
    Invoke \texttt{BandWidSearch}$(\varpi)$ to obtain $\{b_i(\varpi)\}_{\forall i}$\;
    \textbf{if} $\sum\nolimits_{i=1}^I  b_i(\varpi) > B$ \textbf{then} update $\varpi^{\min} = \varpi$ \textbf{else} update $\varpi^{\max} = \varpi$\;
}
Set $\varpi^{\ast} = \varpi$, and recall \texttt{BandWidSearch}$(\varpi^\ast)$ to obtain $\{b_i ^{\ast}\}_{\forall i}$\;
Compute $\{P_i ^{\ast}\}_{\forall i}$ based on Eqn.~(\ref{eq:trans_power})\;
\textbf{return} $\{b_i ^{\ast}, P_i ^{\ast}\}_{\forall i}$\\
\tcc{Function for searching $\{b_i(\varpi)\}_{\forall i}$.}
\textbf{Function} \texttt{BandWidSearch}$(\varpi)$.\\
\For{\normalfont{each device} $i=1,2,\ldots,I$ \textbf{in parallel}}{
Set $b_i^{\max} = B, \forall i$\;
\KwRepeat{$|b_i^{\max} - b_i^{\min}| \leq \varepsilon$}{
    Update $b_i = (b_i^{\max} + b_i^{\min})/2$\;
    Compute $Q(b_i)$ based on Eqn.~(\ref{eq:func-varpi})\;
    \textbf{if} $Q(b_i) +\varpi > 0$ \textbf{then} update $b_i^{\max} = b_i$ \textbf{else} update $b_i^{\min} = b_i$\;
}
Set $b_i(\varpi) = b_i$\;
}
\textbf{return} $\{b_i(\varpi)\}_{\forall i}$\\
\caption{Hierarchical bisection search}
\label{alg:hier-bin-search}
\end{algorithm}

Finally, by first solving Problem (P7) and then putting the optimal $\{b_i^{\ast}\}$ into Eqn.~(\ref{eq:trans_power}), we acquire the optimal solution for Problem (P4). The workflow for solving Problem (P4) is presented in Algorithm \ref{alg:hier-bin-search}, which involves a hierarchical bi-section search to obtain the optimal solution. For the inner search, we aim to solve $ {Q(b_i) + \varpi =0}, \forall i$ and acquire $\{b_i(\varpi)\}_{\forall i}$. For the outer search, we aim to search for $\varpi^\ast$ that satisfies Constraint (\ref{eqn:p5-ctr-3}). Regarding the computational complexity, the number of iteration steps required for inner and outer search are dominated by ${\cal O}\big(\log_2(\max\{B-b_i^{\min}, \forall i\})\big)$ and ${\cal O}\big(\log_2(\varpi^{\max} - \varpi^{\min} )\big)$, respectively. Thus, the overall computational complexity of Algorithm \ref{alg:hier-bin-search} is measured as ${\cal O}\big(\log_2(\varpi^{\max} - \varpi^{\min} )\cdot\log_2(\max\limits_{i}\{B-b_i^{\min}\})\big)$.

\subsection{Cross-Entropy Searching for Problem (P5)}
In this subsection, we aim at solving the top-layer problem that optimizes the time split factor (i.e., $\{\eta _i^\ast\}_{\forall i}$) for each device. However, directly obtaining the closed-form expression for function $E^{\text{round}}\big(\{\eta_i\}_{\forall i}\big)$ is very difficult, hindering the subsequent theoretical analysis. Inspired by \cite{zhu2019novel, sun2016cross}, we employ the learning-based cross-entropy (CE) algorithm as a promising method to address this issue.

To leverage the CE method for solving Problem (P5), we incorporate the optimization variables into a vector denoted as $\bm{\eta}=(\eta _1, \ldots, \eta_I)$. Additionally, we consider $\bm{\eta}\in \mathbb{R}^{I}$ as a random vector, which follows the normal distribution $\bm{\eta} \sim {\cal N}(\bm{\mu}, \bm{\sigma}^2)$. Instead of directly tuning $\bm{\eta}$, we focus on the joint optimization of the vector for mean value $\bm{\mu}=(\mu_1, \ldots, \mu_I)$ and the vector for standard deviation $\bm{\sigma}=(\sigma_1, \ldots, \sigma_I)$.

Let $j$ denote the iteration index and $J$ denote the pre-determined maximum iteration step for the CE algorithm. At the $j$-th iteration, the algorithm samples $M$ feasible solutions $\{\bm{\eta}_{j,m}\}_{\forall m}$ from the distribution profile of $\{\bm{\mu}_j, \bm{\sigma}_j\}$, where $\bm{\eta}_{j,m}\in \mathbb{R}^I$ represents the $m$-th generated solution at the $j$-th iteration. The fundamental concept behind the CE algorithm is to select the top $K$ solutions with the best performance from $\{\bm{\eta}_{j,m}\}_{\forall m}$ for evolving the distribution profile. By iteratively converging towards the optimal distribution with the expectation value of $\bm{\mu}^\ast$, we obtain the numerical result for Problem (P5). The detailed procedure of the CE-based method is presented in Algorithm \ref{alg:ce-search}, which is explained as follows.

\begin{itemize}
    \item (Line \ref{line-alg:init}: Initialization). The algorithm initializes $\bm{\mu}_0=(0.5, \ldots, 0.5)$ and $\bm{\sigma}_0=(1, \ldots, 1)$.
    \item (Line \ref{line-alg:sample}: Sampling). During the $j$-th iteration, the algorithm generates $M$ samples from distribution ${\cal N}(\bm{\mu}_j, \bm{\sigma}_j^2)$ to form the solution set denoted as $\{\bm{\eta}_{j,m}\}_{\forall m}$.
    \item (Line \ref{line-alg:select}: Selection). After invoking Algorithms \ref{alg:bin-search} and \ref{alg:hier-bin-search}, we obtain the corresponding objective values $\{E^{\text{round}}_{j,m}\}_{\forall m}$ for all the generated solutions $\{\bm{\eta}_{j,m}\}_{\forall m}$, where $E^{\text{round}}_{j,m}=E^{\text{round}}(\bm{\eta}_{j,m})$. By sorting $\{E^{\text{round}}_{j,m}\}_{\forall m}$ in an ascending order, we record the first $K$ indices to form a set ${\cal K}\subset \{1,2,\ldots,M\}$. 
    \item (Line \ref{line-alg:update}: Updating). The newly estimated distribution profile parameterized by $\tilde{\bm{\mu}}_{j+1}$ and $\tilde{\bm{\sigma}}_{j+1}$ are respectively updated by
\begin{equation}\label{eq:ce-update}
    \tilde{\bm{\mu}}_{j+1} = \sum \limits _{m\in{\cal K}} \frac{\bm{\mu}_{j,m}}{K}, ~\text{and } \tilde{\bm{\sigma}}_{j+1} = \sum \limits _{m\in{\cal K}} \frac{\bm{\sigma}_{j,m}}{K}.
\end{equation}
    \item (Line \ref{line-alg:smooth}: Smoothness). To reduce fluctuations during iteration, the two parameters of the distribution profile for the next iteration step are respectively smoothed as  
\begin{subequations}\label{eq:ce-smooth}
\begin{empheq}[left=\empheqlbrace]{align}
&\bm{\mu}_{j+1} = \varrho \bm{\mu}_{j+1} + (1- \varrho)\tilde{\bm{\mu}}_{j+1}, \label{eq:ce-smooth-1}\\
&\bm{\sigma}_{j+1} = \varrho \bm{\sigma}_{j+1} + (1- \varrho)\tilde{\bm{\sigma}}_{j+1},\label{eq:ce-smooth-2}
\end{empheq}
\end{subequations}
    where $\varrho \in (0,1)$ is a pre-determined smoothing factor.
\end{itemize}

\begin{algorithm}
\SetAlgoLined
\SetKwInput{KwInput}{Initialization}
\SetKwInput{KwOutput}{Output}
\SetKwRepeat{KwRepeat}{do}{while}
\textbf{Initialization}. Mean value $\bm{\mu}_0=(0.5, \ldots, 0.5)$,  standard deviation $\bm{\sigma}_0=(1, \ldots, 1)$, iteration counter $j=0$, and a small positive number $\varepsilon$.\label{line-alg:init}\\
\While{$j < J  $ \normalfont{and} $ \max\{ \sigma | \sigma \in \bm{\sigma}_{j}\} > \varepsilon$}{
    Generate $M$ samples from ${\cal N}(\bm{\mu}_j, \bm{\sigma}^2_j)$ to form the solutions $\{\bm{\eta}_{j,m}\}_{\forall m}$, and $m\in\{1,2,\ldots, M\}$ \label{line-alg:sample}\;
    Invoke Algorithms \ref{alg:bin-search} and \ref{alg:hier-bin-search} to compute $\{E^{\text{round}}_{j,m}\}_{\forall m}$, where $E^{\text{round}}_{j,m}=E^{\text{round}}(\bm{\eta}_{j,m})$\;
    Select the top $K$ best-performing solutions from $\{\bm{\eta}_{j,m}\}_{\forall m}$, and record the corresponding indices to form ${\cal K}\subset (1, 2, \ldots, M)$ \label{line-alg:select}\;
    Compute $\tilde{\bm{\mu}}_{j+1}$ and $\tilde{\bm{\sigma}}_{j+1}$ according to Eqn.~(\ref{eq:ce-update})\label{line-alg:update}\;
    Smooth the distribution profile to obtain ${\bm{\mu}}_{j+1}$ and ${\bm{\sigma}}_{j+1}$ based on Eqns.~(\ref{eq:ce-smooth-1}) and (\ref{eq:ce-smooth-2})\label{line-alg:smooth}\;
    Update $j=j+1$\;
}
\KwOutput{The converged solution $\bm{\eta}^\ast = \bm{\mu}_j$.}
\caption{Learning-based cross entropy searching}
\label{alg:ce-search}
\end{algorithm}

We now analyze the overall computational complexity for solving Problem (P1). Specifically, there are at most $J$ iteration steps in Algorithm \ref{alg:ce-search}. At each iteration in Algorithm \ref{alg:ce-search}, it invokes $M$ executions for both Algorithms \ref{alg:bin-search} and \ref{alg:hier-bin-search}. We use $\chi$ to denote the maximum search range among the three bi-section searches in Algorithms \ref{alg:bin-search} and \ref{alg:hier-bin-search}, where $\chi$ is computed by
\begin{align}
\small
    \begin{split}
        \chi = \max\big\{\nu^{\max} - \nu^{\min}, \varpi^{\max}-\varpi^{\min}, \max\limits_i\{B-b_i^{\min}\}\big\}.
    \end{split}
\end{align}
The overall computational complexity for obtaining the solution to Problem (P1) can be expressed as
\begin{equation}
    {\cal O}\Big(JM\big(\underbrace{\log_2(\chi)}_{\text{Alg. \ref{alg:bin-search}}}+\underbrace{\log_2^2(\chi)}_{\text{Alg. \ref{alg:hier-bin-search}}}\big)\Big) = {\cal O}\big(JM\log_2^2(\chi)\big).
\end{equation}
Note that the proposed search algorithm can be effectively executed by the server in practical scenarios. The empirical runtime can be disregarded when compared to the latency consumed for model training and parameter transmission.

\subsection{Optimal Augmentation Policy}
Upon obtaining the optimal quantity of synthesized data for each device $\{D_i^{\text{gen}, \ast}\}_{\forall i}$, each device will determine the category-wise data amount locally. Let $C$ denote the total number of label types (i.e., data categories) for the classification task. Each device aims at optimizing the distribution of synthesized data at a category-wise level. 
This involves determining $\{d_{i,c}^{\text{gen}}\}_{\forall c}, (c=1,2,\ldots, C)$ for each individual device.

According to \cite{xu2023blockchain, jiang2023federated}, we adopt the concept of \emph{data entropy} to quantify the distribution of local data. For device $i$, the data entropy of the local mixed dataset ${\cal D}_i^{\text{loc}} \cup {\cal D}_i^{\text{gen}}$ is defined as
\begin{equation}
    H_i = -\sum\limits_{c=1}^C \frac{d_{i,c}^{\text{loc}} + d_{i,c}^{\text{gen}}}{D_{i}^{\text{loc}} + D_{i}^{\text{gen}}}\log_2\Big(\frac{d_{i,c}^{\text{loc}} + d_{i,c}^{\text{gen}}}{D_{i}^{\text{loc}} + D_{i}^{\text{gen}}}\Big).
\end{equation}
where $d^{\text{loc}}_i$ is the data amount of the $c$-th category before the data augmentation at device $i$, and $\sum \nolimits_{c=1}^C d^{\text{loc}}_i = D_{i}^{\text{loc}}$. Intuitively, when the data distribution is uniform across all categories (i.e., $\forall c, d_{i,c}^{\text{loc}} + d_{i,c}^{\text{gen}} = \frac{D_{i}^{\text{loc}} + D_{i}^{\text{gen}}}{C}$), the data entropy attains its maximum value. Conversely, if data is skewed significantly toward a single category (i.e., $\exists c, d_{i,c}^{\text{loc}} + d_{i,c}^{\text{gen}} = {D_{i}^{\text{loc}} + D_{i}^{\text{gen}}}$), the data entropy reaches its minimum value, indicating a strong non-IID distribution.

The augmentation policy optimization at device $i$ is to maximize its local data entropy by regulating the category-wise data amount $\{d_{i,c}^{\text{gen}}\}_{\forall c}$. Formally, the objective of device $i$ is to solve the following optimization problem:
\begin{subequations}\label{eqn:pbm-8}
\begin{align}
    ({\text{P8}}) && &~~~~~\max \limits_{\{d_{i,c}^{\text{gen}} \}_{\forall c}} H_i    &\tag{\ref{eqn:pbm-8}}\\
    {\text{subject to:}} && &~\sum\nolimits_{c=1}^C d_{i,c}^{\text{gen}} = D_i^{\text{gen}} ,&\label{eqn:p8-ctr-1}\\
    {} && &0\le d_{i,c}^{\text{gen}} \le D_i^{\text{gen}} , \forall c. &\label{eqn:p8-ctr-2}
\end{align}
\end{subequations}

\begin{thm}[Optimal local data augmentation]
For a given device $i$,  characterized by its category-wise local dataset distribution $\{d_{i,c}^{\text{loc}}\}_{\forall c}$ and a predetermined data amount of synthesized data $D_i^{\text{gen}}$, the optimal local data augmentation strategy to maximize the data entropy can be expressed as
\begin{align}
\begin{split}
d_{i,c}^{\text{gen}} = \big[\pi_i - d_{i,c}^{\text{loc}}\big]_0^{D^{\text{gen}}_i}.
\end{split}
\label{eq:thm3}
\end{align}
Here, $\pi_i$ represents the variable that needs to be sought in order to satisfy   $\sum\nolimits_{c=1}^C d_{i,c}^{\text{gen}} =  D_i^{\text{gen}}$.
\label{thm:3}
\end{thm}
\begin{proof*}
We first introduce a set of auxiliary variables $\{p_{i,c}\}_{\forall c}$ to denote the category-wise data proportion at device $i$, where $p_{i,c}=\frac{d_{i,c}^{\text{loc}} + d_{i,c}^{\text{gen}}}{D_{i}^{\text{loc}} + D_{i}^{\text{gen}}}$. We can further transform Problem (P8) into the following problem:
\begin{subequations}\label{eqn:pbm-9}
\begin{align}
    ({\text{P9}}) && &\min\limits_{\{p_{i,c} \}_{\forall c}} \sum\limits_{c=1}^C p_{i,c} \log_2(p_{i,c})  &\tag{\ref{eqn:pbm-9}}\\
    {\text{subject to:}} && &~~~\sum\nolimits_{c=1}^C p_{i,c} = 1 ,&\label{eqn:p9-ctr-1}\\
    {} && &p_{i,c}^{\min} \le p_{i,c} \le p_{i,c}^{\max}, \forall c , &\label{eqn:p9-ctr-2}.
\end{align}
\end{subequations}
where $p_{i,c}^{\min}=\frac{d_{i,c}^{\text{loc}}}{D_{i}^{\text{loc}} + D_{i}^{\text{gen}}}$ and $p_{i,c}^{\max}=\frac{d_{i,c}^{\text{loc}} + D_{i}^{\text{gen}}}{D_{i}^{\text{loc}} + D_{i}^{\text{gen}}}$.
It can be verified that Problem (P9) is a convex optimization problem.
By introducing $\{\vartheta_{i,c}\}_{\forall c}$ and $\{\theta_{i,c}\}_{\forall c}$
as the Lagrange multipliers for inequality constraints, and $\varsigma_i$ as the multiplier for the equality constraint.
\begin{subequations}
\begin{empheq}[left=\empheqlbrace]{align}
&\vartheta_{i,c} \ge 0, \; \vartheta_{i,c} (p_{i,c}^{\min} - p_{i,c})=0, \forall c, \label{eqn:p9-kkt-1}\\
&\theta_{i,c} \ge 0, \; \theta_{i,c}(p_{i,c} - p_{i,c}^{\max})=0, \forall c,\label{eqn:p9-kkt-2}\\
& G(p_{i,c}) - \vartheta_{i,c} + \theta_{i,c}  + \varsigma_i = 0, \forall c,\label{eqn:p9-kkt-3}\\
&\text{Constraints (\ref{eqn:p9-ctr-1}) and (\ref{eqn:p9-ctr-2})},
\end{empheq}
\end{subequations}
where $G(p_{i,c})={\log _2}(p_{i,c}) + \frac{1}{{\ln (2)}}$ is an intermediate function.
Furthermore, we have
\begin{subequations}
\begin{empheq}[left=\empheqlbrace]{align}
&\vartheta_{i,c} \ge 0, \; \theta_{i,c} \ge 0, \forall c, \label{eqn:p10-kkt-1}\\
&\big(\underbrace{G(p_{i,c}) + \theta_{i,c}  + \varsigma_i}_{B_1} \big)(p_{i,c}^{\min} - p_{i,c}) = 0, \forall c, \label{eqn:p10-kkt-2}\\
&\big(\underbrace{-G(p_{i,c}) + \vartheta_{i,c}   - \varsigma_i}_{B_2} \big)(p_{i,c} - p_{i,c}^{\max})= 0, \forall c, \label{eqn:p10-kkt-3}
\end{empheq}
\end{subequations}

Next, the solution to Problem (P9) can be divided into the following three cases according to the value of Lagrange multiplier $\varsigma_i$. 
\begin{itemize}
    \item When $\varsigma_i > -{G}(p_{i,c}^{\min})$, we have $B_1>0$. Based on Eqn.~(\ref{eqn:p10-kkt-1}), we obtain $p_{i,c}=p_{i,c}^{\min}$.
    \item When $\varsigma_i < -{G}(p_{i,c}^{\max})$, we have $B_2>0$. Based on Eqn.~(\ref{eqn:p10-kkt-2}), we obtain $p_{i,c}=p_{i,c}^{\max}$.
    \item When $-{G}(p_{i,c}^{\max}) \le \varsigma_i \le -{G}(p_{i,c}^{\min})$, we have $\vartheta_{i,c}=0$ and $ \theta_{i,c}=0$. Thus $\varsigma_i=-{\log _2}(p_{i,c}) - \frac{1}{\ln (2)}$.
\end{itemize}

As a result, the optimal solution to the Problem (P9) is summarized as
\begin{equation}\label{eqn:6-opt}
    {p_{i,c}^\ast} = \Big[ {2^{-(\frac{1}{{\ln (2)}} + \varsigma_i^\ast)}}\Big]_{{p_{i,c}^{\min }}}^{{p_{i,c}^{\max }}}, \;\forall c.
\end{equation}

By putting $p_{i,c}$ into $d_{i,c}^{{\text{gen}}} = {p_{i,c}}(D_i^{{\text{loc}}} + D_i^{{\text{gen}}}) - d_{i,c}^{{\text{loc}}}$, the value of $d_{i,c}^{\text{gen}}$ can be expressed as
\begin{itemize}
    \item When ${p_{i,c}^\ast} = {p_{i,c}^{\min}}$, we have $d_{i,c}^{\text{gen},\ast} = 0$.
    \item When ${p_{i,c}^\ast} = {p_{i,c}^{\max}}$, we have $d_{i,c}^{\text{gen},\ast} = D_i^{\text{gen}}$.
    \item When ${p_{i,c}^\ast} = 2^{-(\frac{1}{{\ln (2)}} + \varsigma_i^\ast)}$, we have $d_{i,c}^{\text{gen},\ast} =  - d_{i,c}^{\text{loc}} + 2^{-(\frac{1}{{\ln (2)}} + \varsigma_i^\ast)}\cdot (D_{i}^{\text{loc}} + D_{i}^{\text{gen}})$.
\end{itemize}
It can be verified that the above solution is equivalent to Eqn.~(\ref{eq:thm3}). Thus, we complete the proof.
\end{proof*}
Note that the numerical value of $\pi _i^{\ast}$ can be efficiently obtained through linear search or bi-section search. Thus, we complete the device-side augmentation strategy optimization.

\section{Experiment Evaluations}
\subsection{Experiment Settings}
\subsubsection{Settings for Edge Devices} 
We consider $I=20$ edge devices for FL training within a 400-meter radius cell, with the base station positioned at the center of the cell. We utilize the path loss model from \cite{yang2020energy} to model the communication link, which is represented as $128.1 + 37.6 \log_{10}(R_i)$ with $R_i$ denoting the distance between the server and device $i$ in kilometers. The power spectral density of additive Gaussian noise is $N_0 = -174$ dBm/MHz and the total bandwidth is $B=20$ MHz. The maximum transmitting power for each device follows the uniform distribution $P_i^{\max} \sim U(20,23)$ dBm. Regarding the model training, the training workload for a single sample is measured as $5\times 10^6$ cycles. The maximum computing frequency and the hardware energy coefficient for each device follow the uniform distributions of $f_i^{\max} \sim U(1,2)$ GHz, and $\epsilon_i \sim U(4\times 10 ^{-27}, 6\times 10 ^{-27})$, respectively.
\begin{figure*}[t]\centering
  \includegraphics[width=0.98\textwidth]{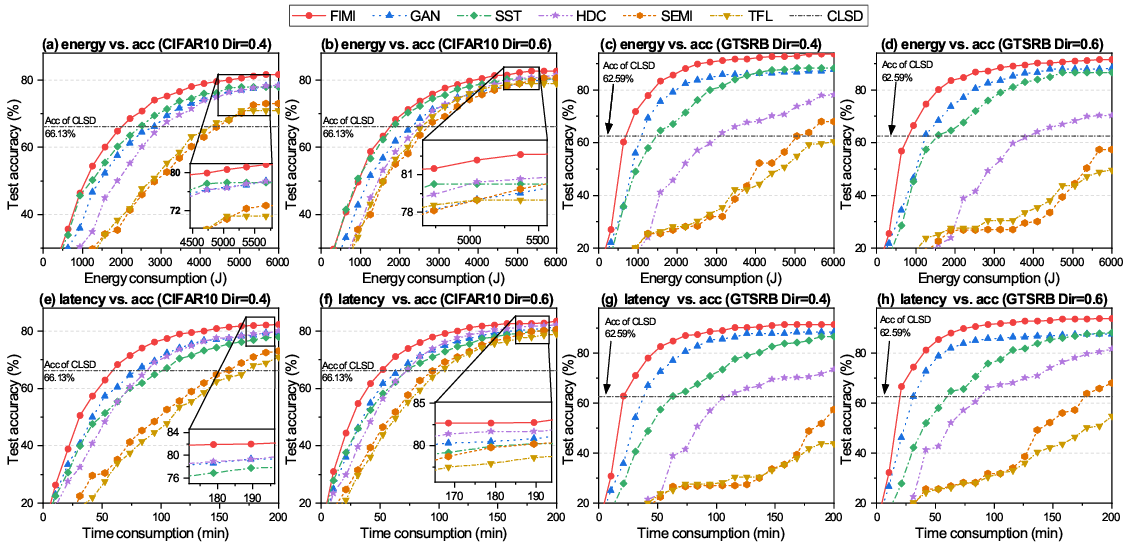}
  \caption{\color{black}Performance comparison of FIMI and baseline methods on CIFAR10 and GTSRB datasets with different levels of non-IID distribution. (The upper row: energy consumption vs. test accuracy; the lower row: time consumption vs. test accuracy.)}\label{fig:exp-comp}
\end{figure*}
\subsubsection{Settings for FL Training} 
Our study is targeted for the implementation of image classification using the CIFAR10 and GTSRB datasets \cite{krizhevsky2009learning, Houben-IJCNN-2013}. We employ the VGG-9 model with the parameter size of 111.7Mb during uplink transmission \cite{simonyan2014very}. 
In terms of data distribution, we evaluate the proposed framework using two Dirichlet distribution levels: Dir(0.4) for skewed non-IID and Dir(0.6) for slight non-IID scenarios \cite{wang2020federated}. For the local dataset, each device is assigned 1250 training samples, i.e., $d_i^{\text{loc}}=1250, \forall i$. For the other hyper-parameters, we set the learning rate as 0.02, the batch size as $B=64$, the local epoch as $\tau=1$, the maximum allowable global error as $\Delta^{\max}=0.2$, the maximum latency per round $T^{\max} = 60$ seconds, and the number of the global rounds as $N=200$.
\subsubsection{Settings for Generative Model}
{\color{black} The pre-trained generative model is trained on the public dataset (i.e., CINIC10) with the classifier-free guidance technique \cite{ho2020denoising, ho2021classifier}. The number of denoising steps is set as 300 and the output image resolution is $32\times32\times3$. Specifically, the total amount of synthesized data is 6659 under the default setting, and it takes about 7.17 minutes for a workstation equipped with eight RTX A5000 GPUs to generate the synthesized data. During the synthesized data distribution, each device receives about 333 samples on average. This yields about 8.1 Mb of downlink traffic per device on average, which is much smaller than 111.7 Mb incurred by the one-time model upload. Therefore, the latency associated with data synthesis and distribution is negligible compared to the numerous rounds of FL training. }

\subsection{Performance Comparisons with Existing Methods}
We conduct performance comparisons between our proposed FIMI approach and several existing methods, including traditional approaches and data augmentation-based FL methods. Their details are as follows. 
\begin{itemize}
	\item {\bf Traditional FL (TFL)}. TFL adheres to the conventional FL paradigm, wherein each device solely uses its local dataset for model training \cite{mcmahan2016comm}.
	\item {\bf Semi-supervised FL (SEMI)}. SEMI incorporates unlabelled samples to enhance model updates \cite{wang2023towards}.    
	\item {\bf Heuristic data compensation (HDC)}. Each device adopts a heuristic scheme, namely, generating synthetic data solely for the category with the least available local data.
    \item  {\color{black}{\bf Server-side training (SST)}. SST preserves the synthesized data at the server, and conducts model training to generate a complementary update every round. The complementary update is aggregated with local model updates to renew the global model.}
    \item {{\bf GAN-based data synthesis (GAN)}. The server uses a pre-trained conditional GAN model to generate additional training samples \cite{xin2020private}.}
    \item {\bf Centralized training with synthesized data (CLSD)}. This method only utilizes the generated synthesized data provided by the server to perform centralized training. 
\end{itemize}

To provide a reasonable comparison,  we adopt the identical optimization algorithm as FIMI for SEMI, HDC, and GAN. In the cases of TFL and SST, the synthesized data of each device is set to zero, and the resource utilization policy is optimized.

Figure \ref{fig:exp-comp} shows the performance comparison between FIMI and baseline methods under different levels of non-IID data distribution. As shown in Figures \ref{fig:exp-comp}(a-d), FIMI outperforms the baseline methods in terms of the test accuracy to reduce energy consumption. Furthermore, Figures \ref{fig:exp-comp}(e-h) show that the proposed FIMI can achieve a faster convergence rate to reduce the training latency.
To evaluate the learning performance, we measure the system cost in terms of energy consumption, training latency, and uplink transmission traffic required to achieve the desired test accuracy for each method. The results are provided in Table \ref{tab:comp}.
{\color{black}The advantages of our proposed FIMI can be summarized into the following three key aspects, i.e., reducing the energy consumption, reducing the time consumption, and improving the converged accuracy.}

\subsubsection{Reducing Energy Consumption} As shown in Table \ref{tab:comp}, our FIMI consumes the energy consumption smaller than the other four methods. Specifically, given the test accuracy of 70\% on the CIFAR10 dataset under Dir(0.4), our FIMI achieves 50\% reduction in the energy consumption compared to TFL. 
{\color{black}As shown in the results, although TFL requires the lowest energy consumption per global iteration, it consumes more iterations than FIMI to achieve the desired accuracy, which consequently degrades the performance of TFL in energy consumption. 
In contrast, FIMI can efficiently achieve the target accuracy with less iterations than TFL. Meanwhile, FIMI reduces the required data size of uplink traffic by approximately 60\% and 50\% compared to TFL under the data distributions of Dir(0.4) and Dir(0.6), respectively. Therefore, our FIMI can effectively reduce the energy consumption than TFL.}

\subsubsection{Reducing Time Consumption} As shown in Table \ref{tab:comp}, our FIMI can effectively reduce the latency for achieving the desired accuracy compared to the other baseline methods. Specifically, when evaluated on the GTSRB dataset with a target accuracy of 80\% under Dir(0.6), FIMI can reduce the latency by 80\% and 70\% compared to HDC and GAN, respectively. {\color{black}It is noticed that the advantage of reduced latency is mainly attributed to the faster convergence rate of FIMI. This is shown in Figures~\ref{fig:exp-comp}(e-h).} Specifically, the convergence rates of the methods ranked according to the descending order are as follows: FIMI, GAN, SST, HDC, SEMI, and TFL.

\begin{table*}
  \caption{Performance comparison between FIMI and other baseline methods.}
  \label{tab:comp}
  \centering
  \setlength{\tabcolsep}{1.0pt}
  \begin{tabular}{llcccccccc}
    \toprule
         & & \multicolumn{4}{c}{\textbf{Dir=0.4} }&\multicolumn{4}{c}{\textbf{Dir=0.6}}\\
    \cmidrule(r){3-6}\cmidrule(r){7-10}
         Dataset & Method & Energy@70 (J)\IEEEauthorrefmark{1} & Latency@70 (h)\IEEEauthorrefmark{1} & Uplink@70 (GB)\IEEEauthorrefmark{1} & Best Acc.(\%) & Energy@78 (J)\IEEEauthorrefmark{1} & Latency@78 (h)\IEEEauthorrefmark{1} & Uplink@78 (GB)\IEEEauthorrefmark{1} & Best Acc.(\%)\\
    \midrule
\multirow{6}{*}{\makecell{\textbf{CIFAR10}}}
& TFL	& 4858.64 (1.0$\times$) 	& 3.22 (1.0$\times$) 	& 56.45 (1.0$\times$) 	& 70.86
& 4506.20 (1.0$\times$) 	& 2.98 (1.0$\times$) 	& 52.36 (1.0$\times$) 	& 78.95\\
& SEMI	& 4913.18 (1.0$\times$) 	& 2.85 (0.9$\times$) 	& 50.02 (0.9$\times$) 	& 73.32
& 4625.86 (1.0$\times$) 	& 2.68 (0.9$\times$) 	& 47.09 (0.9$\times$) 	& 80.54\\
& HDC	& 3507.94 (0.7$\times$) 	& 1.68 (0.5$\times$) 	& 29.54 (0.5$\times$) 	& 79.92
& 4098.38 (0.9$\times$) 	& 1.97 (0.7$\times$) 	& 34.52 (0.7$\times$) 	& 82.05\\
& {\color{black}SST}	& 2844.70 (0.6$\times$) 	& 1.88 (0.6$\times$) 	& 33.05 (0.6$\times$) 	& 77.92
& 3725.80 (0.8$\times$) 	& 2.47 (0.8$\times$) 	& 43.29 (0.8$\times$) 	& 80.23\\
& GAN	& 3403.74 (0.7$\times$) 	& 1.63 (0.5$\times$) 	& 28.66 (0.5$\times$) 	& 79.43
& 4723.56 (1.0$\times$) 	& 2.27 (0.8$\times$) 	& 39.78 (0.8$\times$) 	& 81.34\\
& FIMI	& \textbf{2361.78} (0.5$\times$) 	& \textbf{1.13} (0.4$\times$) 	& \textbf{19.89} (0.4$\times$) 	& \textbf{82.21}
& \textbf{3264.81} (0.7$\times$) 	& \textbf{1.57} (0.5$\times$) 	& \textbf{27.50} (0.5$\times$) 	& \textbf{83.41} \\
\midrule
Dataset & Method & Energy@72 (J)\IEEEauthorrefmark{1} & Latency@72 (h)\IEEEauthorrefmark{1} & Uplink@72 (GB)\IEEEauthorrefmark{1} & Best Acc.(\%) & Energy@80 (J)\IEEEauthorrefmark{1} & Latency@80 (h)\IEEEauthorrefmark{1} & Uplink(GB)\IEEEauthorrefmark{1} & Best Acc.(\%)\\
\midrule
\multirow{6}{*}{\makecell{\textbf{GTSRB}}}
& TFL	& N/A\IEEEauthorrefmark{2} 	& N/A\IEEEauthorrefmark{2} 	& N/A\IEEEauthorrefmark{2} 	& 43.81
&  N/A\IEEEauthorrefmark{2} 	& N/A\IEEEauthorrefmark{2}	& N/A\IEEEauthorrefmark{2}  	& 54.84\\
& SEMI	& N/A\IEEEauthorrefmark{2} 	& N/A\IEEEauthorrefmark{2} 	& N/A\IEEEauthorrefmark{2}	& 57.33
&  N/A\IEEEauthorrefmark{2} 	& N/A\IEEEauthorrefmark{2} 	& N/A\IEEEauthorrefmark{2}  	& 67.93\\
& HDC	& 6807.48 (1.0$\times$) 	& 3.27 (1.0$\times$) 	& 57.33 (1.0$\times$) 	& 73.40
& 6494.89 (1.0$\times$) 	& 3.12 (1.0$\times$) 	& 54.70 (1.0$\times$) 	& 81.59\\
& {\color{black}SST}	& 2467.08 (0.4$\times$) 	& 1.63 (0.5$\times$) 	& 28.66 (0.5$\times$) 	& 86.51
& 2819.52 (0.4$\times$) 	& 1.87 (0.6$\times$) 	& 32.76 (0.6$\times$) 	& 89.04\\
& GAN	& 1806.07 (0.3$\times$) 	& 0.87 (0.3$\times$) 	& 15.21 (0.3$\times$) 	& 88.46
& 2049.19 (0.3$\times$) 	& 0.98 (0.3$\times$) 	& 17.26 (0.3$\times$) 	& 87.69\\
& FIMI	& \textbf{1146.16} (0.2$\times$) 	& \textbf{0.55} (0.2$\times$) 	& \textbf{9.65} (0.2$\times$) 	& \textbf{91.48}
& \textbf{1319.82} (0.2$\times$) 	& \textbf{0.63} (0.2$\times$) 	& \textbf{11.12} (0.2$\times$) 	& \textbf{93.80}\\
    \bottomrule
\multicolumn{10}{l}{\footnotesize{\makecell[l]{\IEEEauthorrefmark{1}``Energy@$x$'', ``Latency@$x$'' and ``Uplink@$x$'' denote the required energy, latency and uplink traffic size to reach $x$\% accuracy, respectively. \\ \IEEEauthorrefmark{2}``N/A'' means that the corresponding method cannot achieve the target global accuracy within the predefined maximum global iteration.}}}
  \end{tabular}
\end{table*}

\subsubsection{Improving Converged Accuracy} As shown in Table \ref{tab:comp}, the proposed FIMI can effectively improve the accuracy of the converged global model, outperforming TFL by 11.35\% and 4.46\% on the CIFAR10 dataset under the Dirichlet data distributions of 0.4 and 0.6, respectively. Meanwhile, we also investigate the performance of model training by solely using AI-synthesized data (i.e., the CLSD method) and show the corresponding results in Figure \ref{fig:exp-comp}. {\color{black}The results show that relying solely on the synthesized data for model training will result in unsatisfactory convergence accuracy, since the AI-generated dataset still shows a different distribution from the local dataset. }
This result highlights the importance of using the generative AI model as a data augmentation method instead of solely using synthesized data for model training.

\begin{figure*}[t]\centering
  \includegraphics[width=0.98\textwidth]{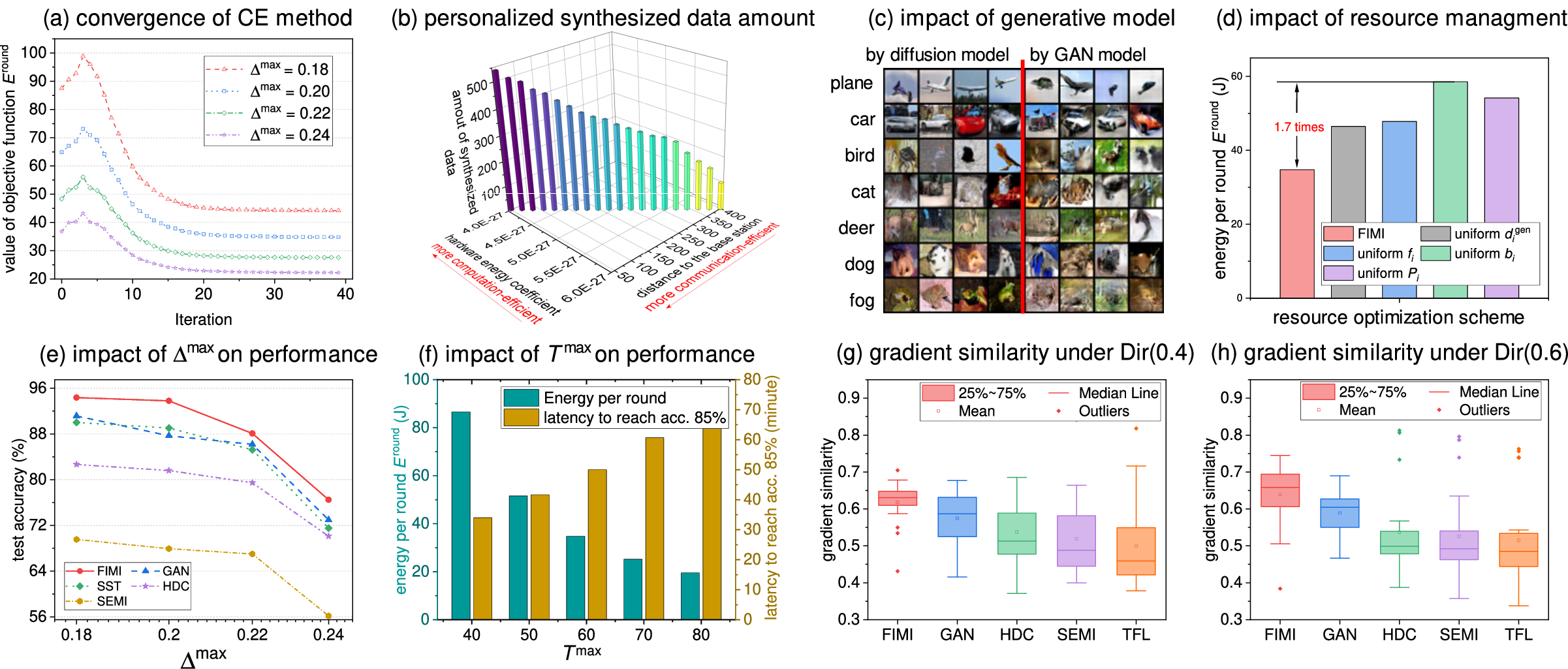}
  \caption{Impact of key mechanisms and hyper-parameters on the generative AI-based FL training system.}\label{fig:abl}
\end{figure*}

\subsection{Detailed Performances of Different Modules in FIMI}
\subsubsection{Performance of CE Method} As shown in Figure~\ref{fig:abl}(a), the proposed CE-based searching algorithm can converge within about 30 iterations. The smaller value of the maximum allowable error $\Delta ^{\max}$ will lead to a higher energy consumption for each round of the global iteration. Moreover, Figure \ref{fig:abl}(b) shows the relationship between the amount of synthesized data $D_i^{\text{gen},\ast}$ with respect to the computation and communication conditions. We construct two arithmetic sequences, one for hardware energy coefficients $\{\epsilon_i\}_{\forall i}$ and the other for the distances $\{R_i\}_{\forall i}$ between the edge devices and the parameter server, and then optimize the data augmentation policy. The result shows that the devices with smaller $\epsilon_i$ and $R_i$ will generate more synthesized data for local training. {\color{black}This result is consistent with the intuition, namely, it will be more beneficial for the devices with favorable channel conditions and energy features to leverage a larger amount of synthesized data for local training.}

\subsubsection{Impact of Generative Models and Hyper-parameters} Figure \ref{fig:abl}(c) shows the synthesized data samples using both the diffusion model and the GAN-based model. {\color{black}The results show that the diffusion-based model can outperform the GAN-based model by generating more realistic images with detailed features.} For example, the synthesized images of the fog produced by the diffusion model exhibit a richer texture than the GAN-based model. The images generated by the GAN-based model appear somewhat blurry. {\color{black}Figure \ref{fig:abl}(d) shows the impact of resource management schemes on the energy consumption. The results demonstrate that the bandwidth allocation will show a significant influence, with a uniform bandwidth allocation policy leading to a 70\% higher energy consumption compared to FIMI.} Figures \ref{fig:abl}(e-f) show the influence of $\Delta^{\max}$ and $T^{\max}$ on the data augmentation-based FL training. We utilize the GTSRB dataset under Dir(0.6) for the experiments. Specifically, a smaller value for $\Delta^{\max}$ yields an improved test accuracy for the converged global model. {\color{black}Our empirical findings suggest that setting the maximum allowable error $\Delta^{\max}$ within the range of $[0.15, 0.25]$ for practical implementations should be appropriate.} Furthermore, the single-round maximum latency $T^{\max}$ regulates the training duration for the FL system. Using a more stringent $T^{\max}$ ( i.e., a smaller value) can reduce the training time, while sacrificing an increased energy consumption.

\subsubsection{Explanation of FIMI} We illustrate the key factors that enable the FIMI to achieve superior performance over existing methods. To shed light on this, we simulate a virtual device holding a locally uniform dataset, resembling IID distribution across all data categories. Let $\bm{g}_0$ denote the local gradient of the virtual device, and let $\bm{g}_i$ denote the local gradients of device $i$ resulting from the mixed dataset. Our goal is to quantify the gradient similarity between the edge device and the virtual device. Based on \cite{zhao2020dataset}, we define the gradient similarity between $\bm{g}_0$ and $\bm{g}_i$ for an $L$-layer shared model as
\begin{equation}
    \texttt{Sim}(\bm{g}_0, \bm{g}_i) = \frac{1}{{2L}}\sum\limits_{l = 1}^L {\bigg(\frac{{ <\bm{g}_0^{[l]},\bm{g}_i^{[l]} > }}{{\| \bm{g}_0^{[l]} \|\| \bm{g}_i^{[l]} \|}} + 1} \bigg).
\end{equation}
Here, $\bm{g}_i^{[l]}$ denotes the gradient vector of $\bm{g}_i$ in the $l$-th layer, $<\cdot, \cdot>$ is the dot product operation, and $\|\cdot\|$ calculates the Euclidean norm for a given vector. As presented in the box plots of Figures \ref{fig:abl}(g-h), the gradient of our FIMI shows the highest level of similarity to that of the IID local data. {\color{black}This result means that our FIMI can effectively mitigate the issue of non-IID data distribution across different edge devices, which thus accelerates the convergence rate of FL training and improves the accuracy of the converged global model.}

\section{Conclusion and Future Work}
In this paper, we have proposed a resource-aware data augmentation framework called FIMI to leverage pre-trained generative AI models to synthesize customized local data for FL training. FIMI can improve the convergence accuracy of FL and save the device-side energy consumption by mitigating data and device resource heterogeneity. Our research results have demonstrated the potential of generative AI in FL and provided the insights for developing efficient and effective FL systems. For our future work, we will explore multi-server empowered AI content generation services in mobile edge computing to facilitate the data synthesis procedure. We will also investigate the incentive mechanisms for the collaborative synergy between generative AI and FL.

\bibliographystyle{IEEEtran}
\bibliography{reference.bib}

\end{document}